\documentclass[10pt,twocolumn,letterpaper]{article}

\usepackage[pagenumbers]{wacv} 

\usepackage{graphicx}
\usepackage{subcaption}
\usepackage[export]{adjustbox}
\usepackage{float}

\definecolor{wacvblue}{rgb}{0.21,0.49,0.74}
\usepackage[pagebackref,breaklinks,colorlinks,allcolors=wacvblue]{hyperref}
\usepackage{subcaption}
\usepackage{dblfloatfix}
\usepackage{booktabs}
\usepackage[table]{xcolor}
\usepackage{longtable}
\usepackage{booktabs}
\def\wacvPaperID{1876} 
\def\confName{WACV}
\def\confYear{2027}

\title{VLMs Win a Systematic Evaluation of Underwater Image Reconstruction}

\author{
Sara Aghajanzadeh \qquad
Yingxue Wang \qquad
Ieva Bagdonaviciute \qquad
David Forsyth\\
University of Illinois Urbana-Champaign\\
{\tt\small saraa5@illinois.edu}
}

\begin{document}

\twocolumn[{
\maketitle

\begin{center}
\captionsetup{type=figure}

\vspace{-0.3cm}

\includegraphics[width=\textwidth]{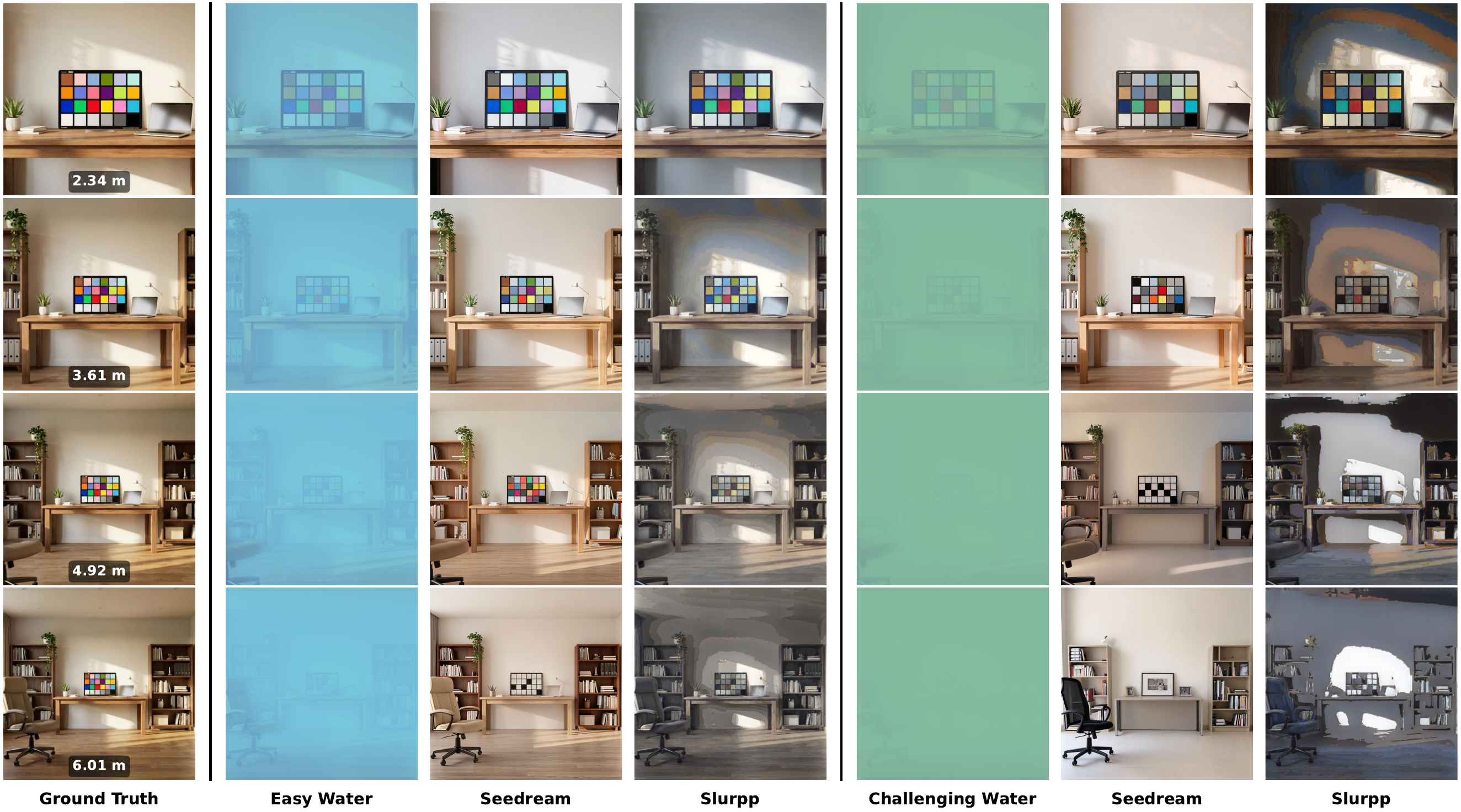}

\vspace{-0.2cm}
\captionof{figure}{ 
      We describe a systematic evaluation pipeline for underwater image restoration models.
      Our pipeline uses images of the same scene viewed from different viewpoints, allowing us to evaluate consistency
      as well as accuracy.  The effects of numerous different water types are then simulated using a
      standard simulation framework~\cite{stsr_uwsim}. A method is evaluated by: restoring each image; comparing results
      to ground truth per image; and comparing consistency across images.   This pipeline allows us to determine the
      effects of changes of depth* and water properties. At short ranges and in clear water, methods produce reasonable reconstructions. As range increases and water quality deteriorates, however, the available scene information rapidly diminishes. Slurpp~\cite{slurpp}, despite being trained with an explicit underwater physics model, struggles to recover the missing information. Surprisingly, Seedream~\cite{gao2025seedream3}, \textit{not} trained with an explicit underwater physics model, 
      relies on its strong image prior to inpaint plausible content.}

\label{fig:teaser}

\end{center}
}]
\footnotetext{*Depth denotes imaging range, i.e., the horizontal distance from the camera to the scene.}
\begin{abstract}
Underwater image restoration consists of recovering an image which looks like
there is no water present.
To date, evaluation has not been systematic.  This paper describes a systematic evaluation pipeline for underwater reconstruction, which can be used to assess a method for accuracy; consistency of reconstruction over camera moves; and  the effect of water parameters.   We use this pipeline to evaluate a range of current procedures, from models constructed using explicit but approximate physical models  of scattering to Vision-Language Models (VLMs which are not currently trained with explicit physical models).
Overall, VLMs  wholly and significantly outperform physically based models in our evaluation, likely because of the importance of a strong image prior.  Results on images of real underwater scenes strongly confirm the evaluation.
\end{abstract}
    
\section{Introduction}
\label{sec:intro}
Water acts like a dense, colored fog that hides important scene details, and restoring underwater images
is now an established problem.  There are numerous recent methods~\cite{watergan,FUnIE-GAN,UIEC$^2$-Net,deepseacolor,pgtie,tctl_net,HCLR-Net}.   It is extremely difficult to evaluate
methods on ground truth, so there are two thrusts in evaluation.
One uses reference-based image quality metrics on synthetic datasets, and no-reference quality metrics or qualitative comparisons on real underwater
images. Some benchmarks also rely on images enhanced by human experts as reference
targets. These evaluations provide useful measures of overall image quality, but there are important missing components.
Qualitative examples can illustrate successes and failures, but they are not enough to understand why a method succeeds or fails.
First is {\em consistency}:  if the camera moves, does the method recover the same color for each shared point?  Second,
there is no systematic evaluation of the effects of water parameters  -- for example, does a method that performs well in
clear water perform badly in turbid water?

We describe an evaluation pipeline built around a carefully designed dataset.  The
dataset is built by constructing images of the same scene from different viewpoints. There are numerous scenes, and each contains a color chart.
Because depth is the key parameter, our images dolly the camera.  Correspondences between
points on the color charts are recorded for each image of every scene. For each image, a standard simulation framework produces underwater
images for many different water types.  The pipeline then reconstructs the original image from each underwater image,
compares reconstructed with true colors, and compares reconstructed colors across images of a scene. 
This procedure systematically varies scene content, imaging depth, and water properties while maintaining known ground
truth.  In turn, evaluation can quantify the effects of depth and water type on any method, compare robustness across
methods, and perform rigorous statistical analyses of restoration behavior under different conditions.

We apply our framework to a range of reconstruction procedures, from those constructed using explicit physical models
to VLMs. Both physical models and VLMs can exhibit nasty behavior.
      When there is very little information in the input, physical models produce huge errors and VLMs produce
      irrelevant content (see bottom-last three images in Fig.~\ref{fig:teaser}).   While VLMs are exposed to no explicit physical model of the effects of water, they strongly outperform
physical methods in all criteria.  We speculate their effectiveness is due to an extremely strong image prior~\cite{vlm_value_learner}.
This suggests that general-purpose visual knowledge learned from large-scale data can be leveraged for other kinds of 
restoration.

The main contributions of this paper are:
\begin{itemize}
    \item A benchmark of RGB–D underwater videos with tracked correspondences, enabling controlled evaluation of restoration consistency across viewpoints, varying depth and water conditions.
    \item A statistical evaluation framework based on linear mixed-effects models for analyzing restoration accuracy, robustness, and color-specific behavior.
    \item A multi-source benchmark spanning five vision-language/video-generation model families, enabling controlled evaluation of potential own-family generation-restoration bias.
    \item A comprehensive evaluation on synthetic and real underwater data showing that modern vision-language models consistently outperform existing underwater restoration methods.
\end{itemize}
\begin{figure*}[t!]
  \centering
  \includegraphics[width=\linewidth]{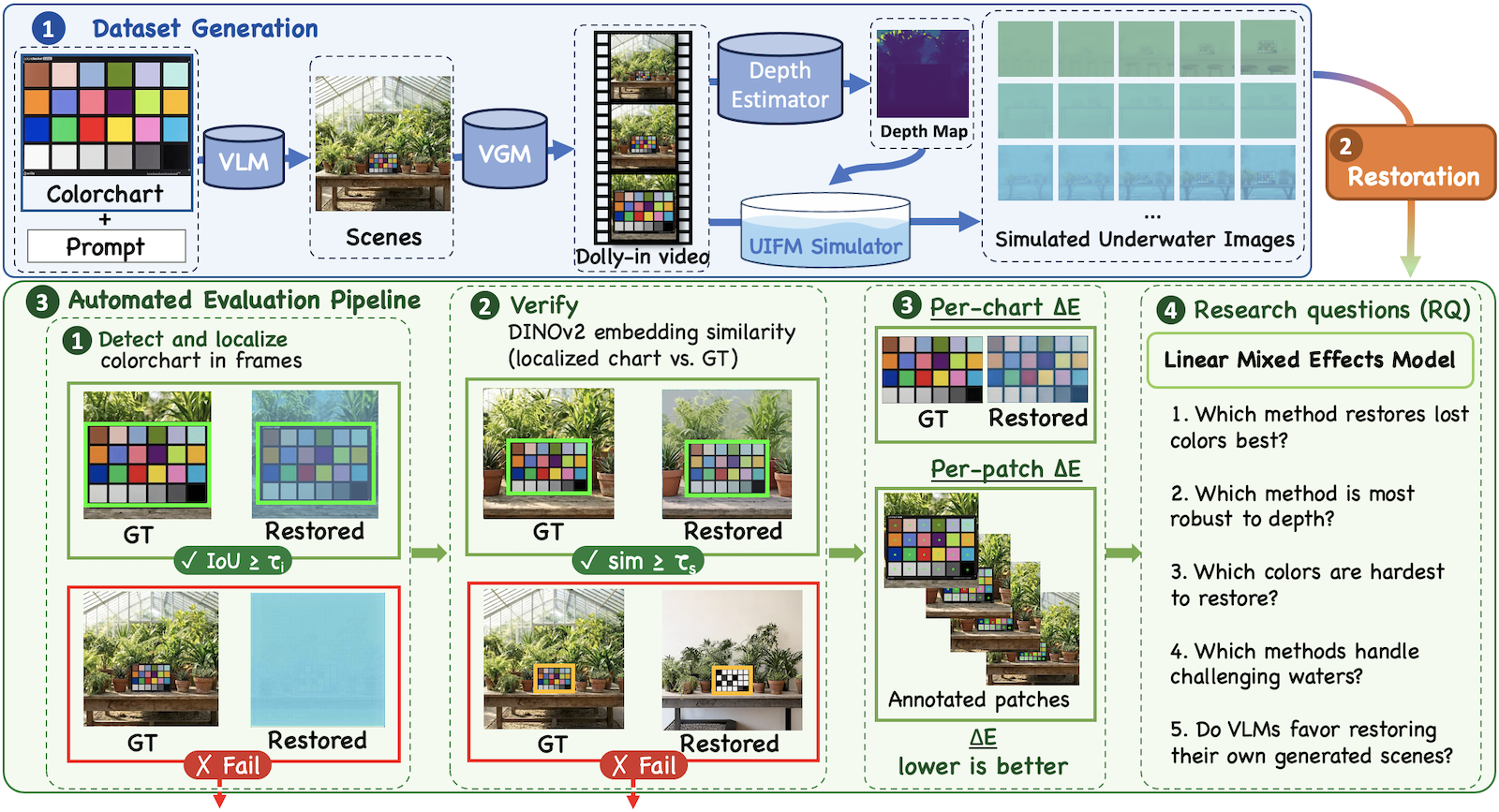}  
  \caption{Our evaluation procedure is built around a dataset ({\bf top}) and a pipeline for evaluating a method ({\bf
      bottom}).  The dataset is built by constructing mutiple videos of a many different static scenes, each containing
    a colorchart, using a dollied camera.  For selected images from this sequence, underwater frames for many different
    water types are constructed using a standard simulator.   Correspondences between
points on the color charts are recorded for each image of every scene.  The pipeline starts with two tests:
does the reconstructed image contain a color chart in the right place?  and is this the right color chart?  For
reconstructions that pass these tests, the color of each patch is compared to ground truth.  Correspondences are used to
evaluate consistency.  Statistical analysis of the errors then yields information about the behavior of the method.}
  \label{fig:approach}
\end{figure*}

\section{Related Work}
\label{sec:related}
\textbf{Water physics models} the observed underwater image \(I\) as the unknown clear scene \(J\) degraded by
attenuation and backscatter: \(I = J e^{-\phi_a D} + \phi^{\infty}\left(1 - e^{-\phi_b D}\right)\), 
where the terms \(e^{-\phi_a D}\) and \(e^{-\phi_b D}\) vary with object range \(D\) and water type~\cite{revised_uifm}.
Changes in depth or water type have strong effects on the image, so our framework explicitly analyzes restoration
performance across varying water types and depths.

\textbf{Datasets.}
Obtaining paired underwater and in-air ground truth data is extremely difficult, and it is fantastically hard to get
paired data for the same scene using different water types. As a result, many datasets rely either
on unpaired real underwater imagery such as Seathru~\cite{seathru} and Squid~\cite{squid_dataset} or on synthetic
degradation pipelines~\cite{slurpp}~\cite{desai_uwsim}. For example, Wu et al.~\cite{slurpp} applies the underwater
image formation model to terrestrial images and estimated depths to simulate diverse underwater data for training
Slurpp, a zero-shot restoration model~\cite{slurpp}. The widely used Underwater Image Enhancement Benchmark
(UIEB)~\cite{uieb_dataset} and recently introduced LSUI~\cite{peng2023ushape} contain real underwater images paired with
human-selected images from enhancement algorithms that can serve as pseudo ground truth. More recently,
Atlantis~\cite{atlantis_dataset} and TIDE~\cite{tide_dataset} use diffusion models to generate synthetic underwater
RGB-D images but lack clear ground-truth references. They also lack realistic scenes, dense depth, camera motion, and
images of the same scene under varying water conditions. Together, these limitations prevent comprehensive restoration
evaluation. 

\textbf{Restoration models.}
Seathru~\cite{seathru} is a physics-based RGB-D method that estimates backscatter, illumination, and attenuation to
recover scene radiance. Deep learning methods instead learn restoration from data. U-shape~\cite{peng2023ushape} uses a
transformer with multi-scale feature fusion, Semi-uir~\cite{huang2023semiuir} adopts semi-supervised learning with
pseudo-labels, and Dtiuie~\cite{lin2026dtiuie} jointly optimizes restoration quality and downstream task
performance. Osmosis~\cite{osmosis} formulates underwater restoration as an RGB-D diffusion inverse problem, combining a
diffusion prior with underwater image formation constraints. 

\textbf{Evaluation protocols.}
Most underwater restoration benchmarks evaluate methods using image-level metrics such as PSNR, SSIM, UIQM~\cite{uiqm}, UCIQE~\cite{uciqe}, NMFC~\cite{nmfc}. While useful, these metrics do not capture physical accuracy and consistency. They tend to favor
images with rich colors ~\cite{niqe,brisque} even when they are not truly faithful
restorations~\cite{Zhang2023UnderwaterIQAColorContrastCIELab}. More broadly, recent computer vision benchmarks
increasingly evaluate models under controlled failure modes rather than relying solely on aggregate
performance~\cite{fire360,robustbench,pai_bench}. We adopt the same philosophy for underwater image restoration. Our evaluation therefore
measures method bias, robustness, and consistency under varying physical conditions. 

\section{Benchmark Dataset}
\label{sec:data}


Our dataset consists of generated dolly-in videos of indoor scenes containing a color chart. Each sequence begins with the camera far from the chart and moves steadily toward it, producing a controlled change in object depth while preserving scene identity. This design enables quantitative evaluation of whether restoration methods recover stable colors for the same physical chart patches as imaging distance changes throughout the sequence. See Fig.~\ref{fig:approach}-(1) for the overview.

\textbf{Scene and video generation.}
We first select a family of vision-language and video-generation models (VGMs). A vision-language model is used to generate a clean terrestrial scene from a text prompt, and a video-generation model from the same family is then used to generate a dolly-in video. We generate data from five different model families to evaluate restoration performance across diverse generative sources and to examine whether restoration models exhibit any bias toward data generated by their own family. Fig.~\ref{fig:restorations} shows this conceptually. Prompts are provided in the supplementary material. Each prompt specifies a realistic indoor environment, the placement and orientation of the color chart, and an initial camera distance of approximately 10--15 meters. We use five arbitrary scenes: an artist studio, a minimalist kitchen, an industrial loft, a home office, and a greenhouse/sunroom. In all scenes, the color chart is constrained to face the camera. For each acceptable starting frame, we generate a slow, steady dolly-in video toward the chart, with prompts explicitly requiring no change in scene composition. This produces clean reference videos with controlled camera motion and a known color target.

We generate clean scenes and subsequently apply underwater degradation using the available simulation framework~\cite{stsr_uwsim}. Our benchmark is agnostic to the specific simulator, maintaining explicit control over the simulation process is important because restoration performance depends strongly on water properties. By varying water conditions in a controlled manner,the benchmark enables practitioners to evaluate restoration methods and select the model that performs best for their target environment.

\textbf{Frame extraction and depth estimation.}
From each generated video, we extract every fifth frame to reduce redundancy while retaining smooth depth variation. For each extracted frame, we estimate dense metric depth using Depth-Anything-3~\cite{da3}. The resulting RGB frames and depth maps are resized to match the input requirements of restoration methods (e.g. \(512 \times 512\) for Slurpp).

\textbf{Underwater simulation.}
Given each clean frame and its estimated metric depth map, we synthesize underwater observations using simulator~\cite{stsr_uwsim}. The degraded image is generated as a function of the clean scene radiance, scene depth, attenuation, backscatter, and veiling light. 
The simulator allows us to include a variety of effects such as forward scattering and spatial variation in water turbidity; our pilot studies in supplementary show no particular change in performance as a result of these effects.
We generate five arbitrary underwater versions of each clean sequence by varying the water parameters. 
Now we have controlled paired data of the form: simulated underwater inputs, corresponding ground truth RGB and metric depth maps, and restoration results. 

\textbf{Colorchart annotations for depth-dependent consistency analysis.}
We'd like to know how a restoration model behaves for the same point in different frames. 
We use a point tracker~\cite{cotracker3} to establish correspondences, then manually verify all tracked annotations.
These are color patch centers on the colorchart in the clean reference sequence. 
Point trajectories are constructed on the clean sequence because the clean video provides more reliable correspondences.
All quantitative analysis is performed on reliable annotations and valid frames. 

\begin{figure}[t]
    \centering
    \includegraphics[width=\linewidth]{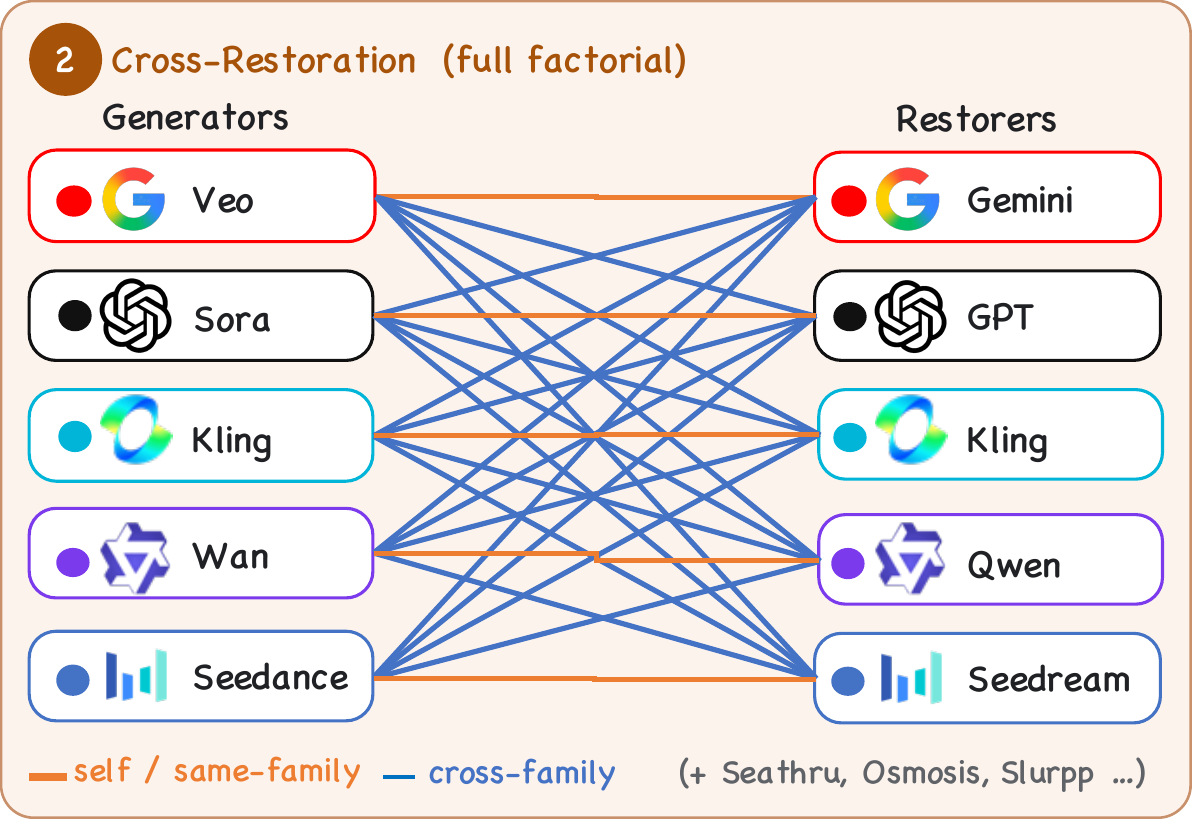}
    \caption{Videos are generated by choosing a family of generative models from the five shown, using the image generator from that family to produce a scene with a color chart, then using the video generator to produce a dollied video of the scene.  This raises the possibillity that reconstruction VLMs might favor evaluation dataset items from their own family in some way.  We investigate this effect by exploring all pairs of (simulating family, reconstructing family) as shown; mostly, there appears to be no effect (Fig.~\ref{fig:rq4_bias_heatmap}).
      }
    \label{fig:restorations}
\end{figure}

\section{Evaluation Framework}
\label{sec:eval}

To evaluate the correctness and consistency of restoration methods on scene and color reconstruction, we use our simulated dataset. This controlled setup allows us to rigorously assess how different methods perform across varying physical conditions, specifically depth, water type, and distinct color patches.

\textbf{Colorchart localization.} The first evaluation step determines whether the colorchart is detected and localized in each restored image. To operate at scale, we use the image-guided OWL-ViT~\cite{owlvit} model with a cropped color-chart image as the visual query. For both the ground-truth and restored images, the model selects the highest-confidence bounding box, and localization accuracy is measured using the Intersection over Union (IoU) metric. A restoration is considered successful if the IoU is at least 0.80; otherwise, it is classified as a failure. This stage serves as a prerequisite for subsequent analysis, ensuring that color-patch evaluations are performed only when the chart has been localized reliably. See Fig.~\ref{fig:approach} (3.1) for a failed example (more examples shown in the supplementary).

\textbf{Colorchart verification.} The second evaluation step verifies that the localized region contains the correct color chart. For each image that passes Step 1, the detected chart regions from the restored and ground-truth images are cropped and embedded using DINOv2~\cite{dinov2}. Their normalized feature embeddings are compared using cosine similarity, and a result is retained when its similarity score is at least 0.50. This step rejects detections that spatially overlap the chart location but do not correspond to the same chart content; an example shown in Fig.~\ref{fig:approach} (3.2).

\textbf{Colorchart fidelity evaluation.}
For restored images that pass the colorchart localization and verification steps, color accuracy is evaluated using $\Delta E_{00}$ metric, where lower values indicate better color preservation. We report two complementary measures. First, a whole-chart evaluation computes the mean pixel-wise $\Delta E_{00}$ between the detected chart regions in the restored and ground-truth images.
Second, a patch-level evaluation computes the same metric within corresponding 8 $\times$ 8 pixel regions centered on the annotated color patches (Fig.~\ref{fig:approach} (3.3)). 
Unless otherwise specified, ground-truth lightness is retained so $\Delta E_{00}$ primarily reflects chromatic differences rather than brightness errors.

\textbf{Statistical evaluation.}
In this study, we seek to answer the five research questions listed in Fig.  \ref{fig:approach} (3.4).

Because the dataset contains repeated observations of the same colorchart and color patches under different depths and water conditions, the measurements are not statistically independent. We therefore use Linear Mixed-Effects Models (LMEMs)~\cite{lmem} to account for these dependencies while estimating the effects of restoration method, depth, water type, dataset, and color patch. Two complementary LMEMs are fitted using $\Delta E_{00}$ as the response variable.

\textbf{Chart-level evaluation.}
The first model uses whole-chart color errors and evaluates overall restoration performance together with method-specific sensitivity to imaging conditions (RQ 1-2,4-5):
\begin{equation}
\begin{aligned}
\Delta E_i =\;&
\beta_0
+ \beta_M \,\mathrm{Method}_i
+ \beta_D \,\mathrm{Depth}_i\\
&+ \beta_W \,\mathrm{Water}_i
+ \beta_S \,\mathrm{Dataset}_i\\
&+ \beta_{MD}\,(\mathrm{Method}_i \times \mathrm{Depth}_i)\\
&+ \beta_{MW}\,(\mathrm{Method}_i \times \mathrm{Water}_i)\\
&+ \beta_{MS}\,(\mathrm{Method}_i \times \mathrm{Dataset}_i)\\
&+ \gamma_{\mathrm{image}(i)}
+ \epsilon_i .
\end{aligned}
\label{eq:chart_lmem}
\end{equation}
In the chart-level model, the method interactions with depth, water type, and dataset quantify how restoration performance changes under these different conditions.

\textbf{Patch-level evaluation.}
The second model uses color errors measured on individual annotated color patches and evaluates color-specific restoration behavior (RQ 3):
\begin{equation}
\begin{aligned}
\Delta E_i =\;&
\beta_0
+ \beta_M \,\mathrm{Method}_i
+ \beta_P \,\mathrm{Patch}_i\\
&+ \beta_D \,\mathrm{Depth}_i
+ \beta_W \,\mathrm{Water}_i
+ \beta_S \,\mathrm{Dataset}_i\\
&+ \beta_{MP}\,(\mathrm{Method}_i \times \mathrm{Patch}_i)\\
&+ \gamma_{\mathrm{image}(i)}
+ \gamma_{\mathrm{patch}(i)}
+ \epsilon_i .
\end{aligned}
\label{eq:patch_lmem}
\end{equation}
In the patch-level model, depth, water type, and dataset are included as adjustment variables, while the method--patch interaction captures color-specific restoration errors. Here, $\beta_0$ denotes the intercept, $\beta$ terms represent fixed-effect coefficients, $\gamma_{\mathrm{image}(i)}$ and $\gamma_{\mathrm{patch}(i)}$ are random intercepts accounting for repeated measurements from the same image and physical patch, respectively, $\epsilon_i$ denotes residual error.

\section{Experimental Results}
\label{sec:res}
\subsection{Experimental Setup}

\textbf{Datasets.}
To evaluate cross-dataset generalization and potential vlm-family-bias, our benchmark consists five different video generation model data: Qwen-set, Seedance-set, Kling-set, Veo-set, and Sora-set. These datasets are generated using Wan~\cite{wan2025}, Seedance~\cite{seedance}, Kling Omni 3.0~\cite{kling}, Veo~\cite{veo}, and Sora~\cite{sora}, respectively. Evaluating restoration methods across all datasets allows us to assess whether models generalize across different synthetic data sources and whether VLM-based restorers exhibit bias toward data generated by their own underlying model family.

\textbf{Models.}
We evaluate eleven models for restoration. These include five SOTA VLMs: Qwen Image Edit~\cite{qwen}, Seedream~\cite{gao2025seedream3}, Kling Omni 3.0~\cite{kling}, Gemini 2.5 Flash Image (``Nano Banana")~\cite{gemini}, and GPT Image 2~\cite{gpt}; six representative and recent underwater image restoration methods: Slurpp~\cite{slurpp} (zero-shot diffusion, TPAMI'25), Osmosis~\cite{osmosis} (zero-shot diffusion, ECCV'24), Dtiuie~\cite{lin2026dtiuie} (CNN+attention, TIP'26), U-shape~\cite{peng2023ushape} (Transformer, TIP'23), Semi-uir~\cite{huang2023semiuir} (Transformer, CVPR'23), and the physics-based method Seathru~\cite{seathru} (CVPR'19). We follow previous work~\cite{atlantis_dataset} and
adopt the unofficial Seathru algorithm implementation. For the rest, official implementations are used without any finetuning.

\textbf{Benchmark and evaluation scale.}
The benchmark contains 25 unique scenes and 125 simulated video sequences spanning five different waters. For each sequence, we evaluate 5--10 frames sampled at different imaging depths. Each frame contains a 24-patch color chart used for quantitative analysis.
Across the five datasets, the evaluation framework processes restorations corresponding to
\(5\ \text{datasets} \times 25\ \text{scenes} \times 5\text{--}10\ \text{frames} \times 11\ \text{restoration methods}.\)
For patch-level analyses, each valid restoration contributes measurements from 24 color-chart patches. Because restorations that fail the chart localization or semantic validity stages are excluded from subsequent analyses, the effective sample size is reduced to several thousand valid restorations. Exact sample counts for each evaluation stage are reported in supplementary.


\textbf{Additionally}, we evaluate a subset of real images containing colorcharts from the Squid dataset~\cite{squid_dataset}. We also evaluate Kling Omni 3.0 on the simulated Osmosis benchmark and compare it with the quantitative results reported by Osmosis~\cite{osmosis} using brightness-invariant PSNR, SSIM, and LPIPS.

\begin{figure}[t]
    \centering
    \includegraphics[width=\linewidth]{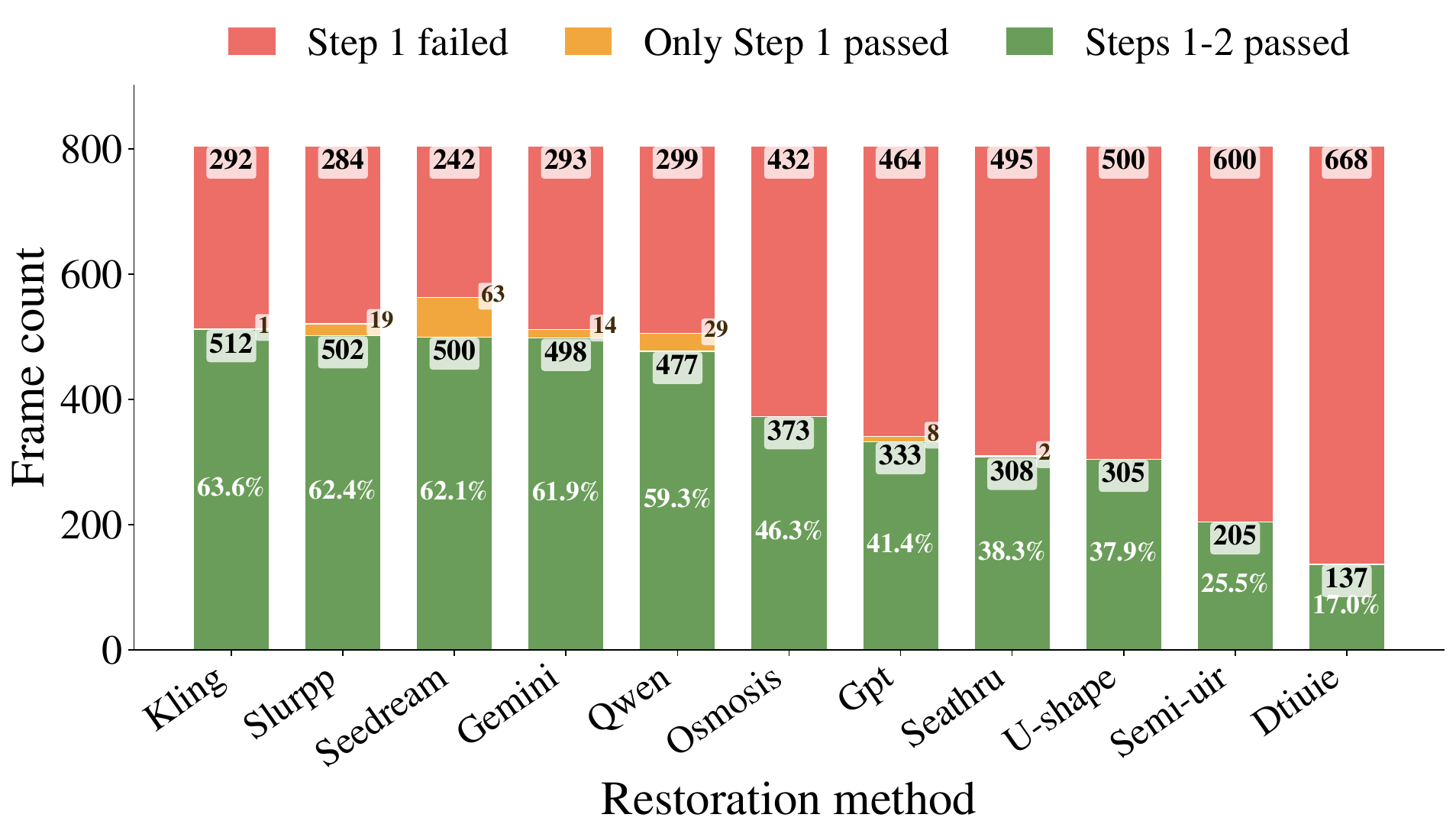}
    \caption{\textbf{Colorchart localization (Step 1) and verification (Step 2).} Each bar summarizes the 805 expected frames for a restoration method. Step 1 requires IoU between localized and ground-truth charts $\geq\tau_{\mathrm{IoU}}=0.8$ and Step 2 requires DINOv2 similarity $\geq\tau_{\mathrm{sim}}=0.5$. Methods are ordered by Step-2 success rate. Percentages denote the fraction of frames passing both steps.}
    \label{fig:step1-step2-summary}
\end{figure}

\subsection{Quantitative Results}
\textbf{Colorchart localization.}
Step 1 measures whether restored images preserve sufficient chart structure for successful localization. Fig.~\ref{fig:step1-step2-summary} shows that VLMs consistently achieve higher localization success rates than specialized underwater restoration methods 
(failures: chart missing or distorted).

\textbf{Feature-level colorchart verification.}
Step 2 measures whether successfully localized charts preserve their visual appearance using the cosine similarity between DINOv2 embeddings of the localized and ground-truth charts. As shown in Fig.~\ref{fig:step1-step2-summary}, success rates decrease for nearly all methods relative to Step 1.

\begin{figure}[]
    \centering
    \begin{subfigure}{\linewidth}
        \centering \includegraphics[width=\linewidth]{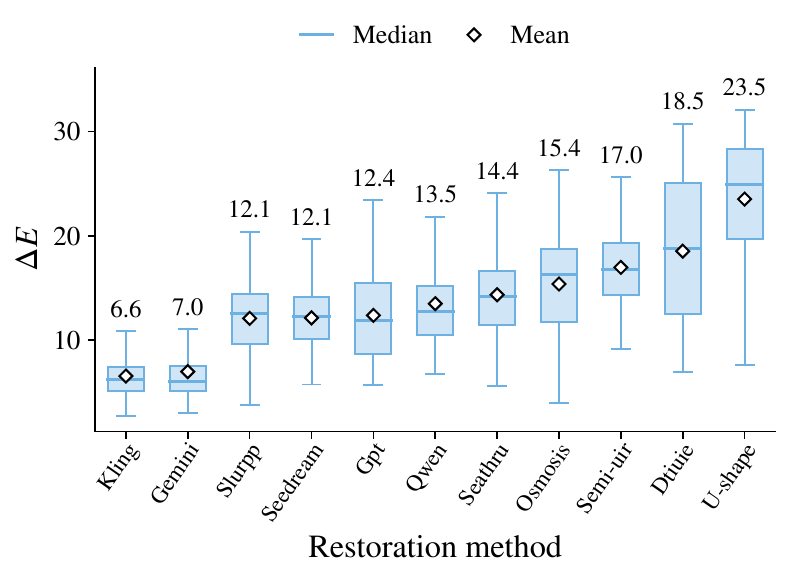}
        \caption{Distribution of color-chart $\Delta E$ scores across restoration methods.}
        \label{fig:deltae-boxplot}
    \end{subfigure}
    \begin{subfigure}{\linewidth}
        \centering
        \includegraphics[width=\linewidth]{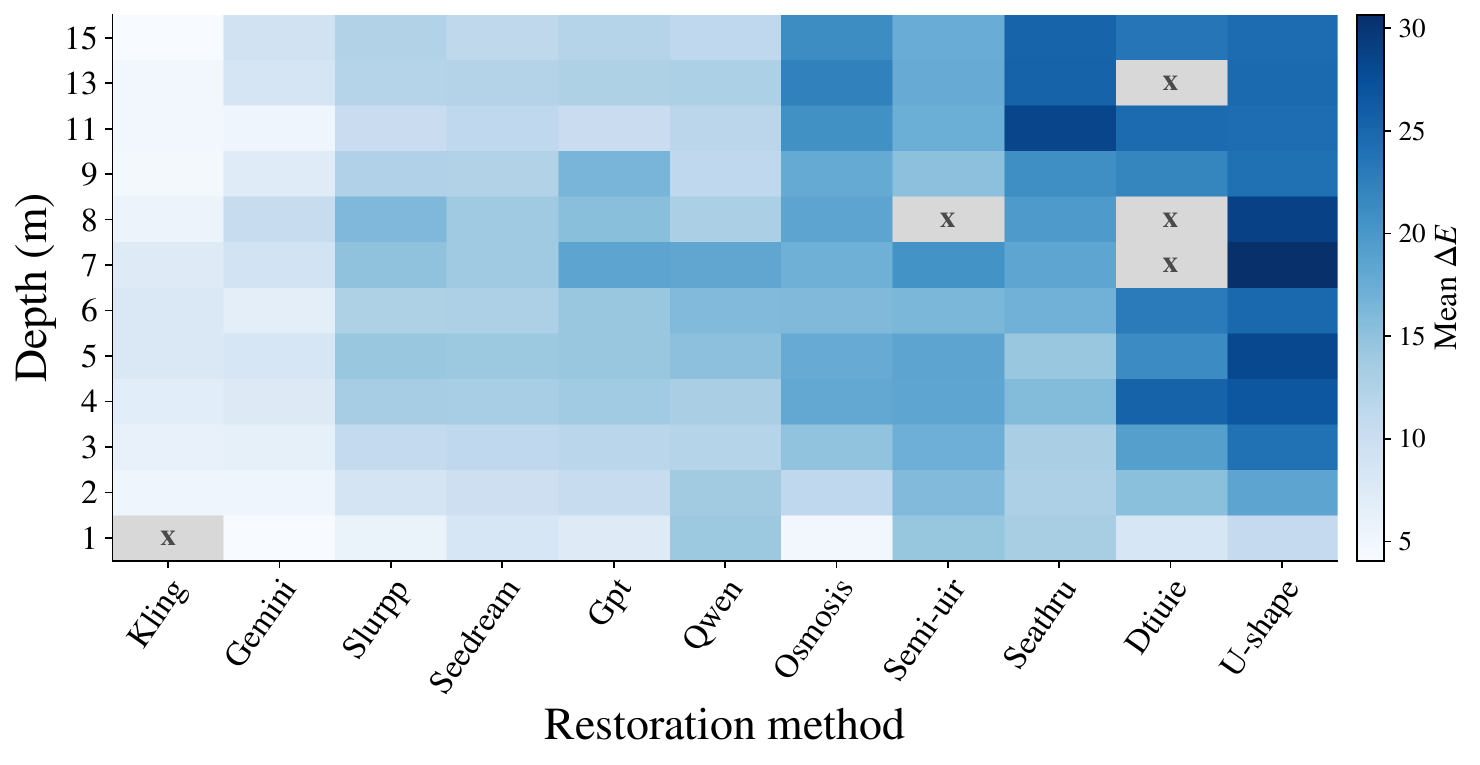}
        \caption{Mean color-chart $\Delta E$ by method and its range from the camera}
        \label{fig:deltae-depth-heatmap}
    \end{subfigure}
    \caption{\textbf{Colorchart fidelity.} Accuracy of restored colorcharts. Lower $\Delta E$ indicates better agreement with the ground truth. (a) Distribution of $\Delta E$ by method (means shown above boxplots). (b) Mean $\Delta E$ across depth bins. Kling and Gemini perform on par and ahead of the others. X indicates no restorations.}
    \label{fig:deltae-color-accuracy}
\end{figure}

\textbf{Colorchart fidelity.}
Among successfully verified charts, Kling and Gemini achieve the lowest color errors ($\Delta E$), while traditional underwater restoration methods perform substantially worse (Fig.~\ref{fig:deltae-boxplot}). Fig.~\ref{fig:deltae-depth-heatmap} further shows that Kling and Gemini remain robust with increasing depth, whereas competing methods become progressively less accurate.

\textbf{RQ1: Which method restores lost colors best?}

We use the fitted model to estimate the expected color restoration error \(\Delta_E\) for each method while accounting for variation due to depth, water type, dataset, and repeated observations. The model predictions are reported at a common reference condition (water type I, Veo-set, average depth = 3.51 m), allowing methods to be compared under identical conditions. The choice of reference conditions is described in the supplementary material.
Tab.~\ref{tab:rq1_adjusted_accuracy} summarizes the model-predicted chart-level restoration errors. Gemini achieves the lowest predicted error ($\Delta E=4.91$), indicating the most accurate color restoration overall. Kling's not statistically significantly different from Gemini ($p=0.125$), suggesting that the two methods perform comparably. All remaining methods exhibit significantly larger errors ($p<0.001$).

\begin{table}
\centering
\caption{Predicted chart-level restoration errors at the reference condition (water type I, Veo-set, mean depth = 3.51 m). Lower $\Delta E$ is better. Increase is relative to Gemini. Depth slope gives the predicted increase in $\Delta E$ per additional 1 m of imaging depth (depth-sensitivity curves shown in Supplementary). $p$-values below 0.001 indicate statistical significance. Kling is statistically on par with Gemini ($p=0.125$).}
\label{tab:rq1_adjusted_accuracy}
\begin{tabular}{lrrrl}
\toprule
Method & $\Delta E$ & Increase & Depth slope & $p$-value \\
\midrule
Gemini & 4.91 & reference & +0.50 & reference \\
Kling & 5.65 & \textcolor{red}{+0.75} & +0.24 & 0.125 \\
Slurpp & 9.87 & \textcolor{red}{+4.96} & +0.75 & $<0.001$ \\
Seedream & 10.99 & \textcolor{red}{+6.08} & +0.58 & $<0.001$ \\
Gpt & 11.17 & \textcolor{red}{+6.27} & +0.68 & $<0.001$ \\
Qwen & 11.40 & \textcolor{red}{+6.50} & +0.48 & $<0.001$ \\
Osmosis & 13.57 & \textcolor{red}{+8.66} & +1.26 & $<0.001$ \\
Seathru & 14.13 & \textcolor{red}{+9.22} & +1.37 & $<0.001$ \\
Semi-uir & 18.18 & \textcolor{red}{+13.27} & +0.49 & $<0.001$ \\
Dtiuie & 20.75 & \textcolor{red}{+15.85} & +2.40 & $<0.001$ \\
U-shape & 22.65 & \textcolor{red}{+17.74} & +1.47 & $<0.001$ \\
\bottomrule
\end{tabular}
\end{table}

\begin{figure*}[!t]
    \centering
    \includegraphics[width=\textwidth]{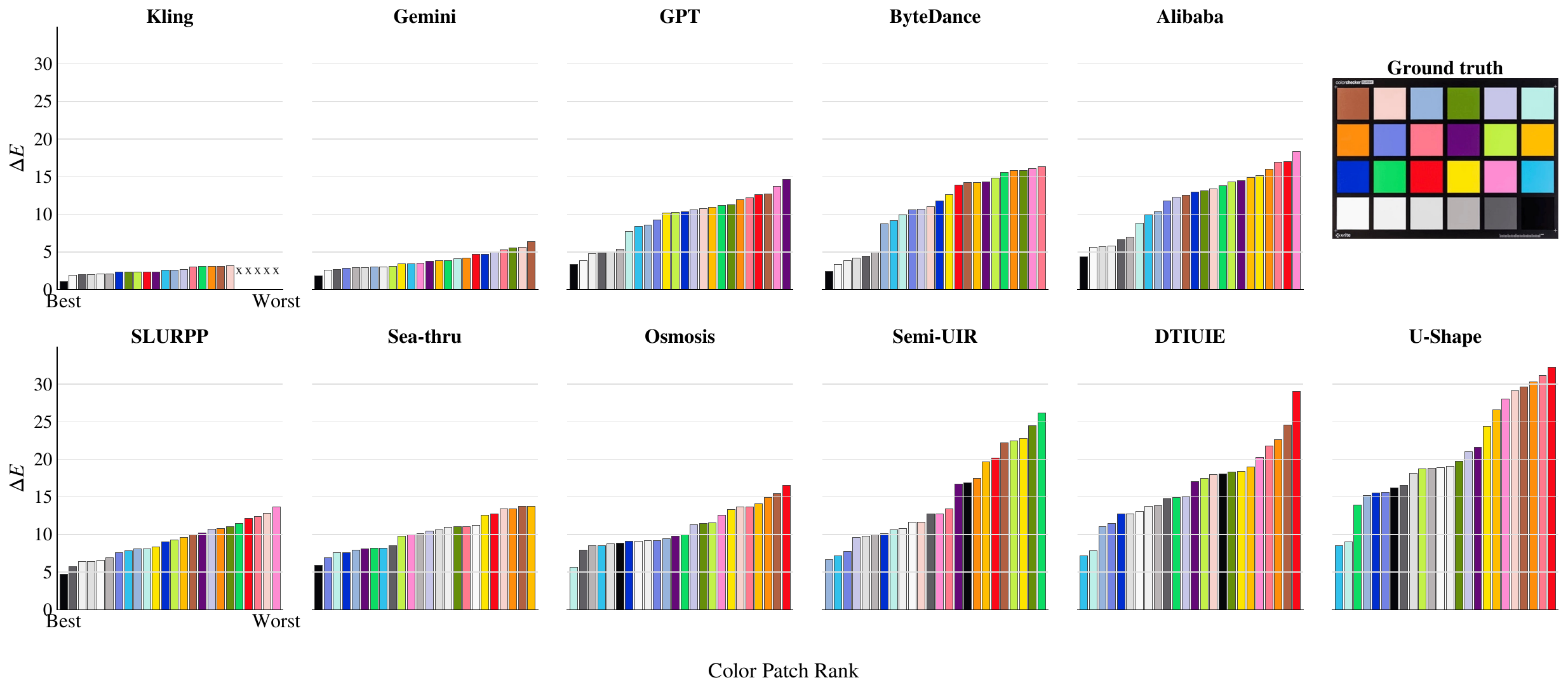}
    \caption{Within-method ranked color-patch errors. Each subplot ranks color patches from lowest predicted $\Delta E$ on the left (best) to highest predicted $\Delta E$ on the right (worst), with bars colored by the ground-truth colorchart swatch. \textbf{Top} row shows VLMs and \textbf{bottom} row shows UIR methods. X for Kling means for a few colors, the predicted error was not significantly different than reference baseline Gemini's error (\(p>0.001\)) hence not included.}
    \label{fig:rq3_patch_ranked_bars}
\end{figure*}

\textbf{RQ2: which method is most robust to depth?}

The model tests whether restoration error changes with depth at different rates across methods via the method \(\times\) depth interaction. The interaction was significant (\(p<0.001\)), indicating different depth-error slopes. As shown in Tab.~\ref{tab:rq1_adjusted_accuracy}, Gemini has a slope of +0.50 $\Delta E$ per depth unit (1m). Kling shows the shallowest degradation (+0.24), while Dtiuie shows the steepest (+2.40). Thus, larger positive slopes indicate faster degradation with depth, whereas smaller slopes indicate greater robustness.


\textbf{RQ3: Which colors are hardest to restore?}

Fig.~\ref{fig:rq3_patch_ranked_bars} shows color patch rankings for each restoration method. Kling achieves the lowest mean significant patch error (2.43 $\Delta E$), followed by Gemini (3.84), whereas U-Shape exhibits the highest mean significant patch error (20.76). Across methods, saturated warm colors (red, orange, yellow, and some green) are consistently the hardest to restore, whereas neutral, blue, and gray patches are the easiest, consistent with stronger attenuation of longer wavelengths underwater.

\textbf{RQ4: Which methods handle challenging waters?}

The method \(\times\) water interaction tests whether methods respond differently to water types and was significant (\(p<0.001\)). Water types are ranked from I (easiest) to V (hardest) using a controlled visibility test (Supplementary). Kling has the lowest mean error across water types (6.30 $\Delta E$), followed by Gemini (6.39). Kling is also the least water-sensitive (range 1.11 $\Delta E$), whereas Semi-uir is the most water-sensitive (range 3.94 $\Delta E$). Thus, restoration accuracy depends on both water type and restoration method.
\begin{figure}[b]
    \centering
    \includegraphics[width=\linewidth]{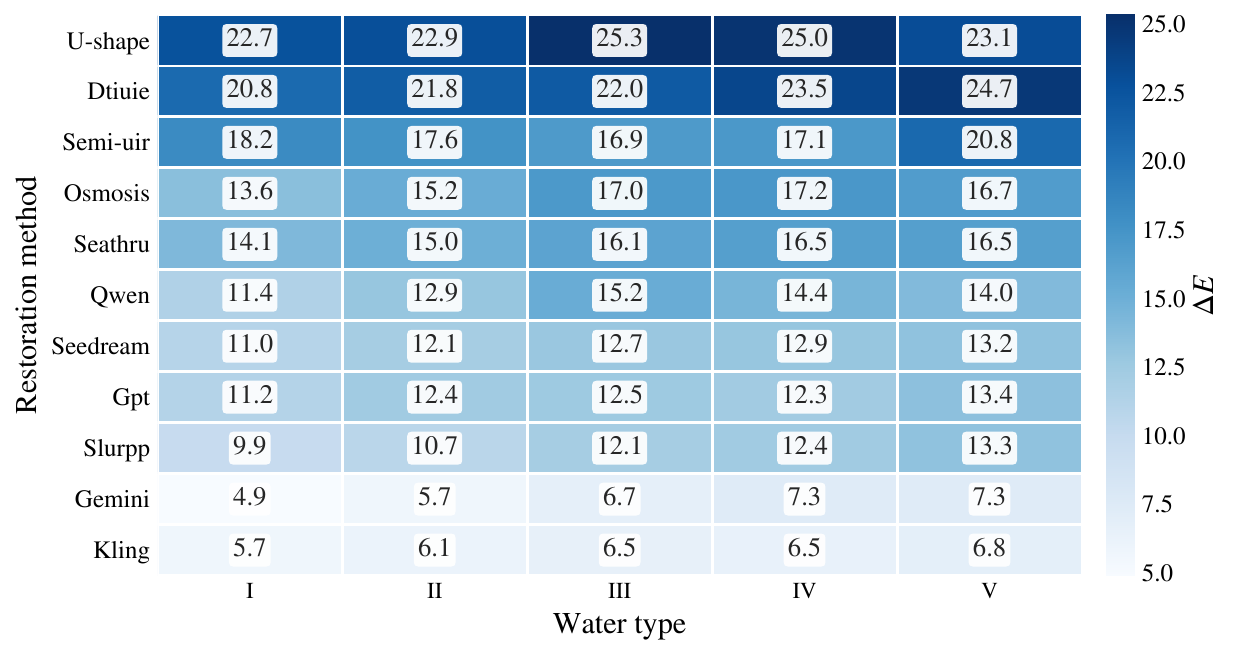}
    \caption{\textbf{Model-predicted chart-level color restoration error by method and water type.} Water types are ordered from easiest (I) to hardest (V). Darker cells indicate higher $\Delta E$. Kling is the preferred choice when the water type is unknown, whereas Gemini is preferred for water types I-II.}
    \label{fig:water_sensitivity}
\end{figure}

\begin{figure*}[!t]
    \centering
    \includegraphics[width=\textwidth]{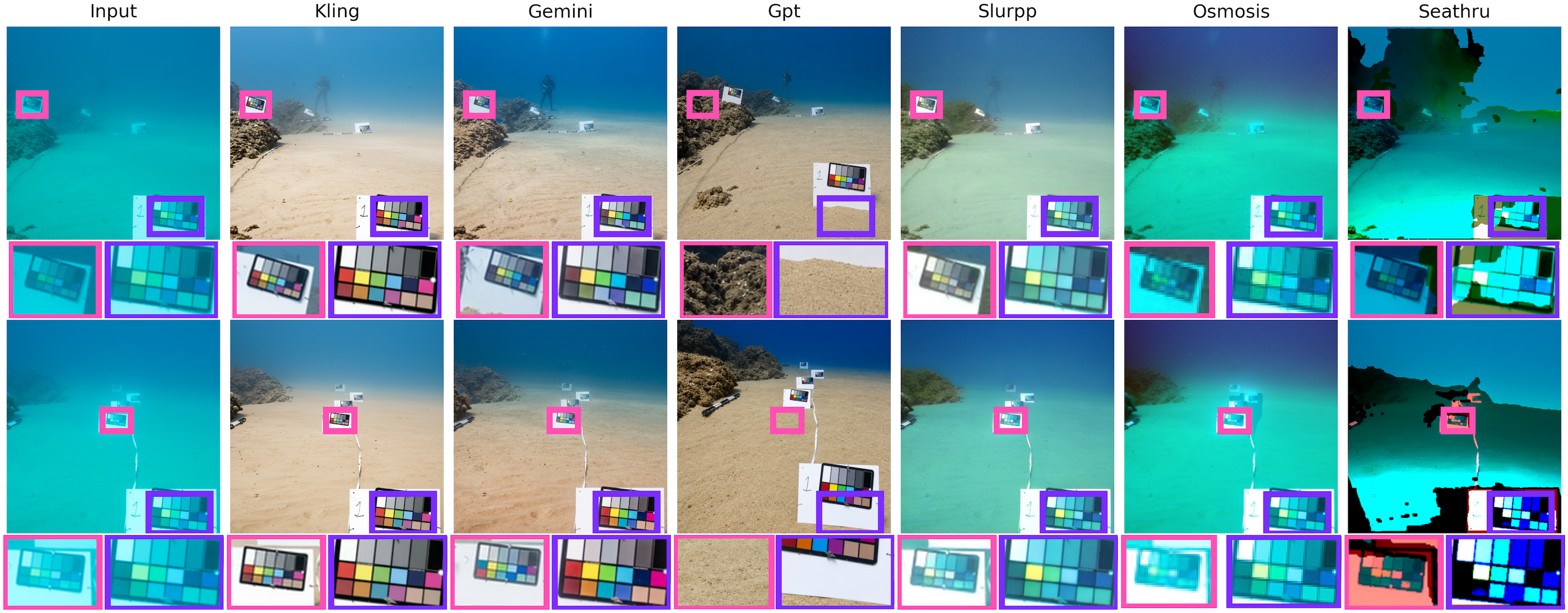}
    \caption{Real underwater images from Squid dataset~\cite{squid_dataset} restored by 3 VLMs (Kling, Gemini, Gpt), 2 zero-shot models (Slurpp, Osmosis), and 1 physics-based method (Seathru).
      Note that comparing color charts at front and back yields an assessment of consistency.  Because the same chart is used in these images,
      consistency can be assessed by comparing the far and near charts (detail in the {\bf bottom} row). Gemini has produced quite different charts -- comparing the two
      charts in the bottom scene with the nearest chart in the top scenes suggests that three entirely different charts  were used, which is not correct.
      VLMs can produce reconstructions that are spatially not equivalent to the input image -- here Gpt has distorted the scene (compare color chart locations in {\bf pink} and {\bf purple}). Slurpp and Osmosis have very odd reconstructions, and Seathru was unable to reconstruct satisfactorily.}
    \label{fig:real_restorations}
\end{figure*}

\begin{figure}[b]
\centering
\includegraphics[width=\linewidth]{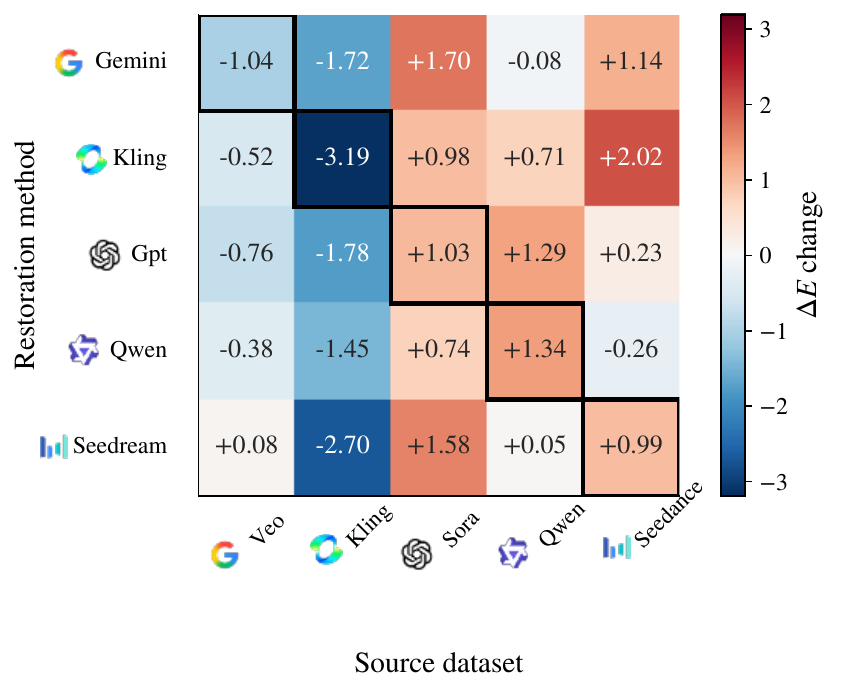}
\caption{VLM-family bias analysis. Heatmap of predicted colorchart errors across methods and dataset sources. Kling shows a clear preference for its own generated dataset, Gemini a weak preference, and the remaining VLMs none. The method-by-dataset interaction is significant ($p<0.001$).}
\label{fig:rq4_bias_heatmap}
\end{figure}
\textbf{RQ5: Do VLMs favor restoring their own generated scenes?} 

The LMEM tests whether dataset effects differ across VLM restorers via the method \(\times\) dataset interaction, which was significant (\(p<0.001\)). The predictions show a clear own-dataset advantage for Kling, a weaker advantage for Gemini, and no advantage for Seedream, Gpt, or Qwen. Thus, own-dataset bias is not uniform across VLM families.

\textbf{Evaluation on the Osmosis benchmark.}
We additionally evaluate Kling Omni 3.0 on the simulated Osmosis benchmark using the original evaluation protocol. As shown in Tab.~\ref{tab:osmosis_comparison}, Kling achieves the best PSNR and SSIM and competitive LPIPS~\cite{lpips}, outperforming dedicated underwater restoration methods despite being a general-purpose vision-language image editing model.

Real underwater images are qualitatively evaluated in Fig.~\ref{fig:real_restorations}, with additional results in the supplementary. Kling produces the strongest restorations. Gemini occasionally recovers incorrect colors, while Gpt introduces spatial shifts. Slurpp and Osmosis leave substantial water degradations, and Seathru does not reconstruct satisfactorily.

\begin{table}[]
\centering
\caption{Comparison on the simulated Osmosis benchmark. Results for baslines are reproduced from~\cite{osmosis}. Higher PSNR and SSIM are better, while lower LPIPS is better.}
\label{tab:osmosis_comparison}
\begin{tabular}{lccc}
\toprule
Method & PSNR $\uparrow$ & SSIM $\uparrow$ & LPIPS $\downarrow$ \\
\midrule
Contrast stretch & 17.13 & 0.83 & 0.11 \\
Semi-uir~\cite{huang2023semiuir} & 17.82 & 0.83 & 0.12 \\
Ucolor~\cite{ucolor} & 17.92 & 0.83 & 0.10 \\
Usuir~\cite{usuir} & 16.76 & 0.80 & 0.18 \\
UW-Net~\cite{uwnet} & 18.04 & 0.75 & 0.26 \\
WaterNet~\cite{waternet} & 17.27 & 0.82 & 0.11 \\
Unveiling~\cite{unveiling} & 16.34 & 0.79 & 0.18 \\
Osmosis~\cite{osmosis} & 22.74 & 0.89 & \textbf{0.06}\\
\midrule
\textbf{Kling Omni 3.0 (ours)} & \textbf{25.75} & \textbf{0.88} & 0.10 \\
\bottomrule
\end{tabular}
\end{table}

\section{Conclusion}
We presented a systematic evaluation framework for underwater image restoration that measures reconstruction accuracy and robustness to changing conditions. Our results show that modern vision-language models consistently outperform state-of-the-art physics-based restoration methods on both simulated and real underwater data, highlighting the value of strong image priors. Future work includes improving foundation models for underwater restoration through prompt optimization, test-time adaptation, and lightweight domain-specific tuning.


{
    \small
    \bibliographystyle{ieeenat_fullname}
    \bibliography{main}
}
\clearpage
\setcounter{page}{1}
\maketitlesupplementary

\setcounter{figure}{0}
\renewcommand{\thefigure}{S\arabic{figure}}

\setcounter{table}{0}
\renewcommand{\thetable}{S\arabic{table}}

\section*{Supplementary Note 1}

\textbf{Generation prompts.}
We generated clean terrestrial scenes using Gemini Nano Banana and then generated corresponding dolly-in videos using Veo. Similar prompts were used for the other model families.

All scene prompts specified that:
(i) the provided color chart should be reproduced exactly,
(ii) the color chart should face the camera,
(iii) the initial camera should be positioned approximately 10--15\,m from the chart.

The scene-specific prompts were:

\begin{itemize}
    \item \textbf{Artist's Studio:} The color chart leaning against a clean white canvas on a wooden easel in a bright studio with neutral gray walls and high-contrast lighting.

    \item \textbf{Minimalist Kitchen:} The color chart resting on a kitchen island in a chic minimalist kitchen containing a coffee machine and flowers.

    \item \textbf{Industrial Loft:} The color chart placed on a polished concrete floor inside a high-ceiling loft with large windows.

    \item \textbf{Home Office:} The color chart placed beside a silver laptop on a desk.

    \item \textbf{Greenhouse/Sunroom:} The color chart positioned on a table surrounded by several green potted plants, with its size adjusted naturally to fit the scene.
\end{itemize}

For every scene, the corresponding Veo prompt was:

\begin{quote}
\emph{Generate a slow, steady dolly-in toward the color chart. Keep the scene composition unchanged throughout the video. Use the provided scene image as the starting frame.}
\end{quote}

For Veo generation, the synthesized scene image served as the first frame of the video, while the ground-truth color chart image was provided as the final frame to guide the camera motion toward the chart.

\newpage
\section*{Supplementary Note 2}

\textbf{Restoration prompts.}
All vision-language models are prompted identically for restoration on the simulated benchmark. The prompt is:

\begin{quote}
\small
You are an expert underwater image enhancement model.

Given an underwater image, generate a reconstructed image of the same scene as if it were captured in air (no water), with natural colors, correct contrast, no haze or backscatter, no changes to the scene composition.

Return the reconstructed image.
\end{quote}

For the real underwater image experiments, we additionally provided the ground-truth color chart as a reference image, since this improves restoration quality. Accordingly, Kling, Gemini, and GPT receive the same prompt with the following additional instruction:
\begin{quote}
\small
If there are color charts in the image, restore them to have colors similar to those shown in the provided ground-truth color chart image.
\end{quote}

\newpage

\begin{figure*}[!t]
    \centering
    \includegraphics[width=0.8\textwidth]{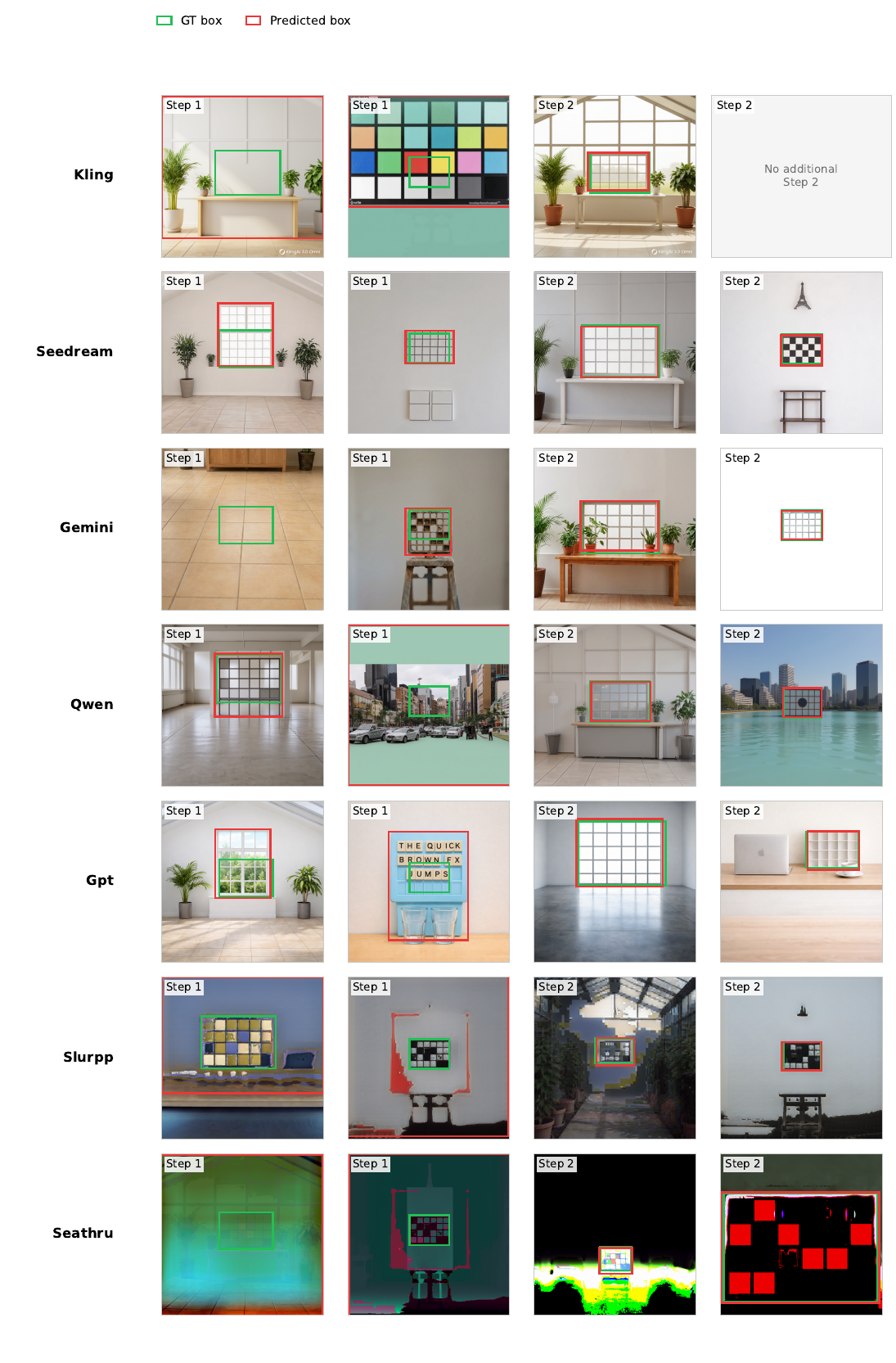}
    \caption{\textbf{Supplementary Note 3.} Two failure examples shown per step 1: chart detection and localization and step 2: chart verification. Green bounding box shows where the colorchart is in ground truth images, and red is the predicted box in the restored images.}
    \label{fig:step1-2fails}
\end{figure*}

\clearpage
\section*{Supplementary Note 4}
\textbf{Forward scattering and inhomogeneous water effects.}
\paragraph{Setup.} Like many underwater simulation pipelines, our main benchmark synthesizes degraded observations with the revised underwater image formation model (UIFM), which models direct attenuation, backscatter, and veiling light but assumes spatially homogeneous water and neglects forward scattering. To probe the sensitivity of our evaluation to these simplifications, we conducted a pilot study on the Veo-set in which we augmented the standard simulation with two additional effects: (i) \emph{forward scattering}, modeled as a depth-dependent blur applied to the direct-transmission component, and (ii) \emph{spatially inhomogeneous attenuation}, modeled as a Gaussian-random-field modulation of the per-channel attenuation coefficients~\cite{stsr}. Both additions make the simulation more physically realistic, since in real underwater imaging light is not only absorbed and backscattered but also scattered in the forward direction by suspended particles, while water properties are rarely uniform across the scene due to spatial variations in particle concentration, illumination, and medium composition. For every restoration method, we recomputed chart $\Delta E_{00}$ averaged over all scenes, water types, and evaluation frames, and compared it against the standard UIFM baseline.

\begin{figure*}[]
    \centering
    \begin{subfigure}[t]{0.49\linewidth}
        \centering
        \includegraphics[width=\linewidth]{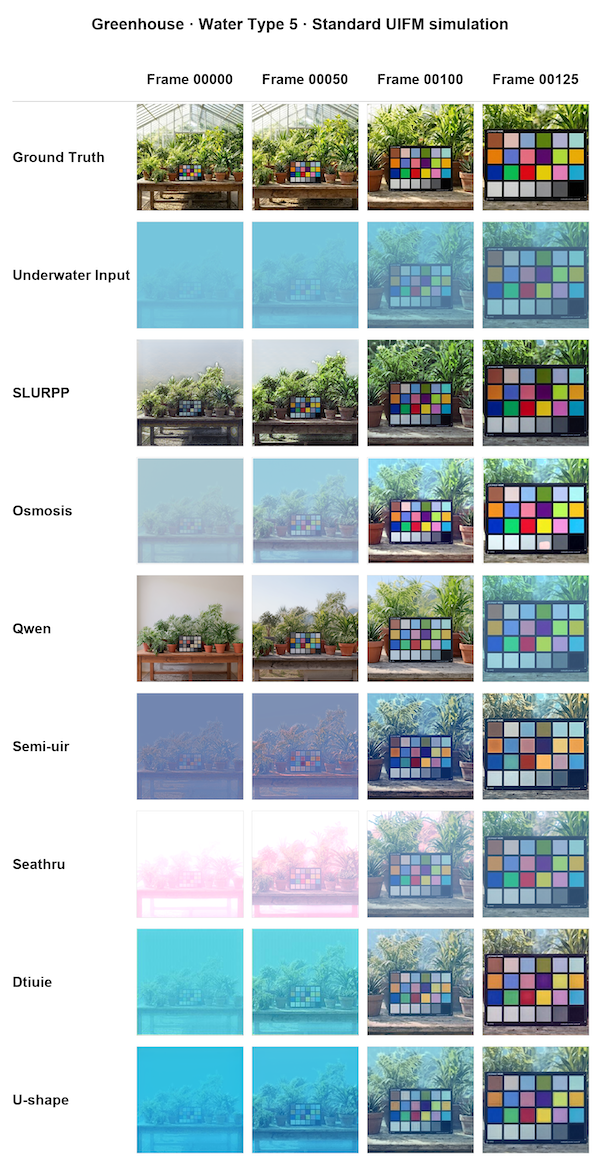}
        \caption{Standard UIFM simulation.}
        \label{fig:qual_std_uifm}
    \end{subfigure}\hfill
    \begin{subfigure}[t]{0.49\linewidth}
        \centering
        \includegraphics[width=\linewidth]{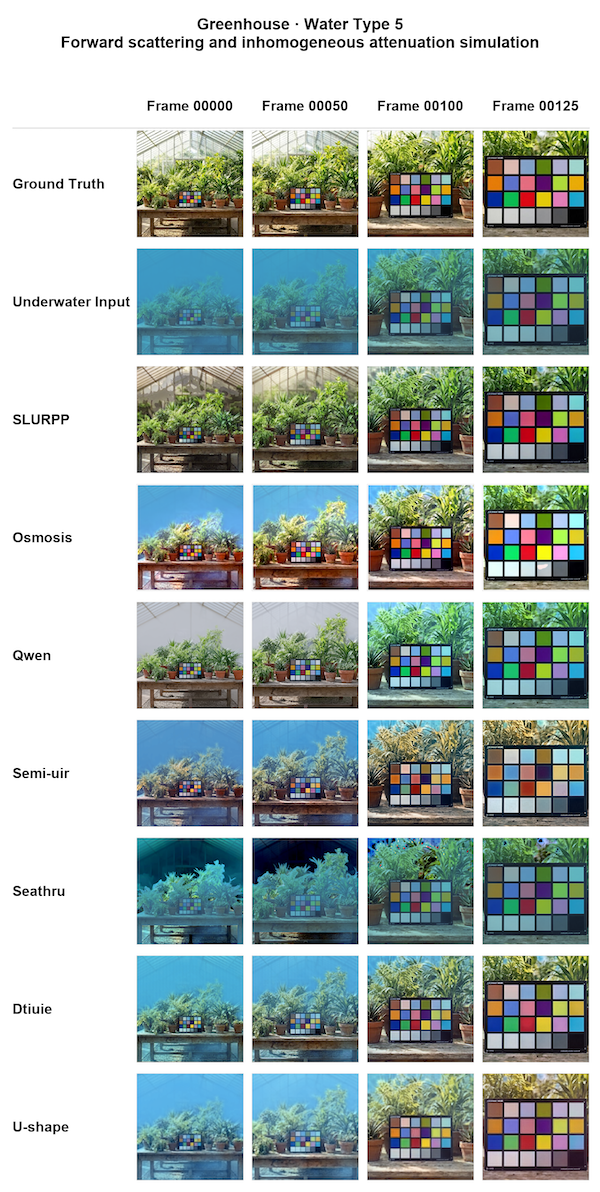}
        \caption{Forward-scattering + inhomogeneous attenuation simulation.}
        \label{fig:qual_fscatter_inhom}
    \end{subfigure}
    \caption{\textbf{Qualitative comparison under two simulation models} on the same scene, water type, and frame set. The left panel uses the standard UIFM simulation, while the right panel uses the extended simulation with forward scattering and inhomogeneous attenuation.}
    \label{fig:qual_two_models}
\end{figure*}

\paragraph{Results.} Table~\ref{tab:fwdscatter} reports the results. Adding forward scattering and inhomogeneous attenuation consistently reduces mean chart $\Delta E_{00}$ for all seven methods. Absolute improvements range from $-3.80$ (Slurpp) to $-9.81$ (U-Shape). This indicates that, in our pilot setting, making the simulation closer to real underwater physics actually leads to easier restoration and better color recovery. We attribute this trend to two complementary effects: first, forward scattering acts as a low-pass filter on the direct signal, suppressing high-frequency distortions and making the degradation smoother; second, although the attenuation field becomes spatially varying, its local fluctuations are largely averaged out over the small color-chart patches used for evaluation, so it does not introduce strong patch-level instability. At the same time, the relative ranking is not strictly preserved: while Slurpp and Qwen remain the top two methods, mid-tier swaps appear (e.g., Osmosis vs.\ Semi-uir, and Dtiuie vs.\ U-shape), indicating method-dependent sensitivity to the updated physics. Qualitative comparisons are shown in Fig.~\ref{fig:qual_two_models}, which visually corroborate the quantitative trend in Table~\ref{tab:fwdscatter}.

\paragraph{Limitations.} This pilot study should be interpreted with some caution. The forward-scattering kernel and the random-field parameters were chosen as plausible approximations rather than being calibrated from in-situ optical measurements, so they should not be regarded as a fully faithful physical model of any specific underwater environment. In addition, the observed reduction in $\Delta E_{00}$ does not necessarily imply that every aspect of the restoration problem becomes easier in a perceptual sense; rather, it shows that under our chart-based color evaluation, the added physical realism produces degradations that are more favorable for color recovery. A more complete study would require calibrated measurements of real water optical properties and broader validation on additional scenes and tasks, which we leave for future work.

\begin{table}[H]
\centering
\caption{Forward-scattering and inhomogeneous-water pilot study (Veo-set).
Mean chart $\Delta E_{00}$ (lower is better) under the standard UIFM versus
the extended model with forward scattering and spatial inhomogeneity.}
\label{tab:fwdscatter}
\begin{tabular}{lccc}
\toprule
Method & \shortstack{Standard\\UIFM} & \shortstack{+ Fwd.\ scatter\\+ inhomog.} & $\Delta$ \\
\midrule
Slurpp   & 12.86 & 9.06  & $-3.80$ \\
Seathru  & 18.41 & 13.84 & $-4.58$ \\
Osmosis  & 15.69 & 10.85 & $-4.85$ \\
Qwen     & 16.43 & 10.73 & $-5.69$ \\
Semi-UIR & 18.30 & 12.57 & $-5.73$ \\
DTIUIE   & 23.60 & 13.95 & $-9.66$ \\
U-Shape  & 25.47 & 15.66 & $-9.81$ \\
\bottomrule
\end{tabular}
\end{table}

\clearpage

\clearpage
\begin{table*}[t]
\centering
\small
\setlength{\tabcolsep}{4pt}
\renewcommand{\arraystretch}{1.15}
\caption{Sample sizes corresponding to heatmap fig.~\ref{fig:deltae-depth-heatmap} in the main paper}
\label{tab:deltae-depth-bin-counts}
\begin{tabular}{l|rrrrrrrrrrr}
\toprule
\rowcolor{gray!15}
\textbf{Depth upper (m)} & \textbf{Kling} & \textbf{Gemini} & \textbf{Slurpp} & \textbf{Seedream} & \textbf{Gpt} & \textbf{Qwen} & \textbf{Osmosis} & \textbf{Semi-uir} & \textbf{Seathru} & \textbf{Dtiuie} & \textbf{U-shape} \\
\midrule
15 & 5 & 5 & 5 & 4 & 3 & 3 & 1 & 1 & 1 & 1 & 2 \\
13 & 5 & 5 & 5 & 5 & 3 & 2 & 2 & 1 & 1 & 0 & 3 \\
11 & 4 & 4 & 5 & 5 & 4 & 3 & 3 & 1 & 2 & 1 & 2 \\
9 & 10 & 9 & 12 & 10 & 6 & 8 & 5 & 3 & 6 & 1 & 5 \\
8 & 12 & 9 & 21 & 19 & 6 & 7 & 10 & 0 & 8 & 0 & 1 \\
7 & 36 & 27 & 18 & 25 & 4 & 29 & 21 & 1 & 2 & 0 & 1 \\
6 & 31 & 22 & 22 & 24 & 14 & 21 & 9 & 7 & 11 & 5 & 12 \\
5 & 96 & 84 & 84 & 77 & 46 & 76 & 56 & 13 & 40 & 7 & 30 \\
4 & 87 & 91 & 91 & 92 & 46 & 93 & 60 & 27 & 56 & 16 & 45 \\
3 & 160 & 168 & 169 & 166 & 133 & 163 & 139 & 94 & 126 & 57 & 137 \\
2 & 66 & 69 & 65 & 68 & 63 & 67 & 62 & 52 & 50 & 44 & 62 \\
1 & 0 & 5 & 5 & 5 & 5 & 5 & 5 & 5 & 5 & 5 & 5 \\
\rowcolor{orange!14}
\textbf{Total} & \textbf{512} & \textbf{498} & \textbf{502} & \textbf{500} & \textbf{333} & \textbf{477} & \textbf{373} & \textbf{205} & \textbf{308} & \textbf{137} & \textbf{305} \\
\bottomrule
\end{tabular}
\end{table*}

\section*{Supplementary Note 5}
\begin{figure}[]
    \centering
    \includegraphics[width=\linewidth]{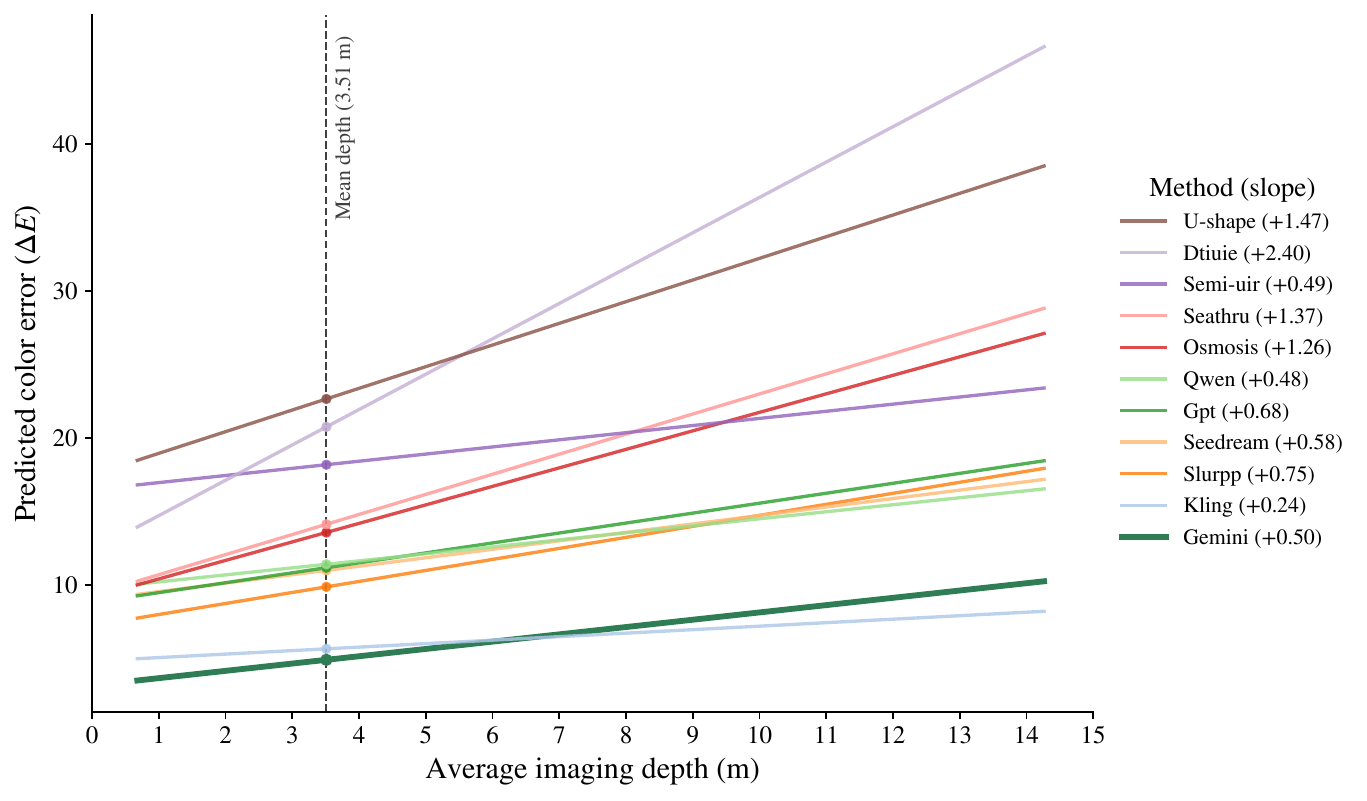}
    \caption{\textbf{Predicted depth-sensitivity slopes} for chart-level color restoration error across the observed imaging-depth range (0.68-14.26 m). Slope is the expected Delta E change per additional depth unit. Steeper positive slopes indicate faster degradation with depth.}
    \label{fig:depth_slope}
\end{figure}

\begin{figure}[]
    \centering
    \includegraphics[width=0.95\linewidth]{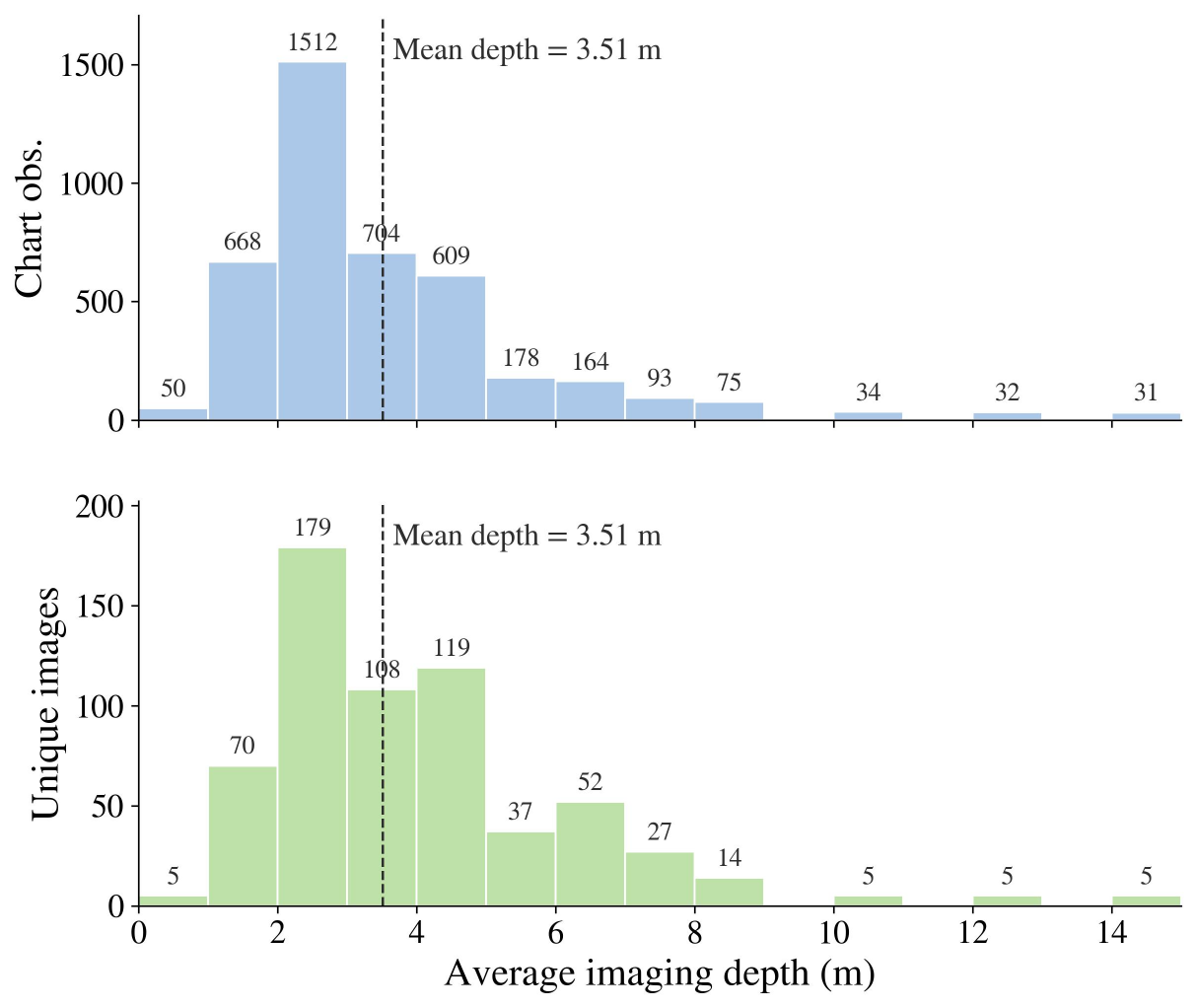}
    \caption{Histogram coverage of depths in the benchmark (i.e., sample size for fitting the model to estimate depth sensitivity slopes shown in Fig.~\ref{fig:depth_slope}).}
    \label{fig:histogram_depth_coverage}
\end{figure}

\begin{figure*}[!t]
    \centering
    \includegraphics[width=\textwidth]{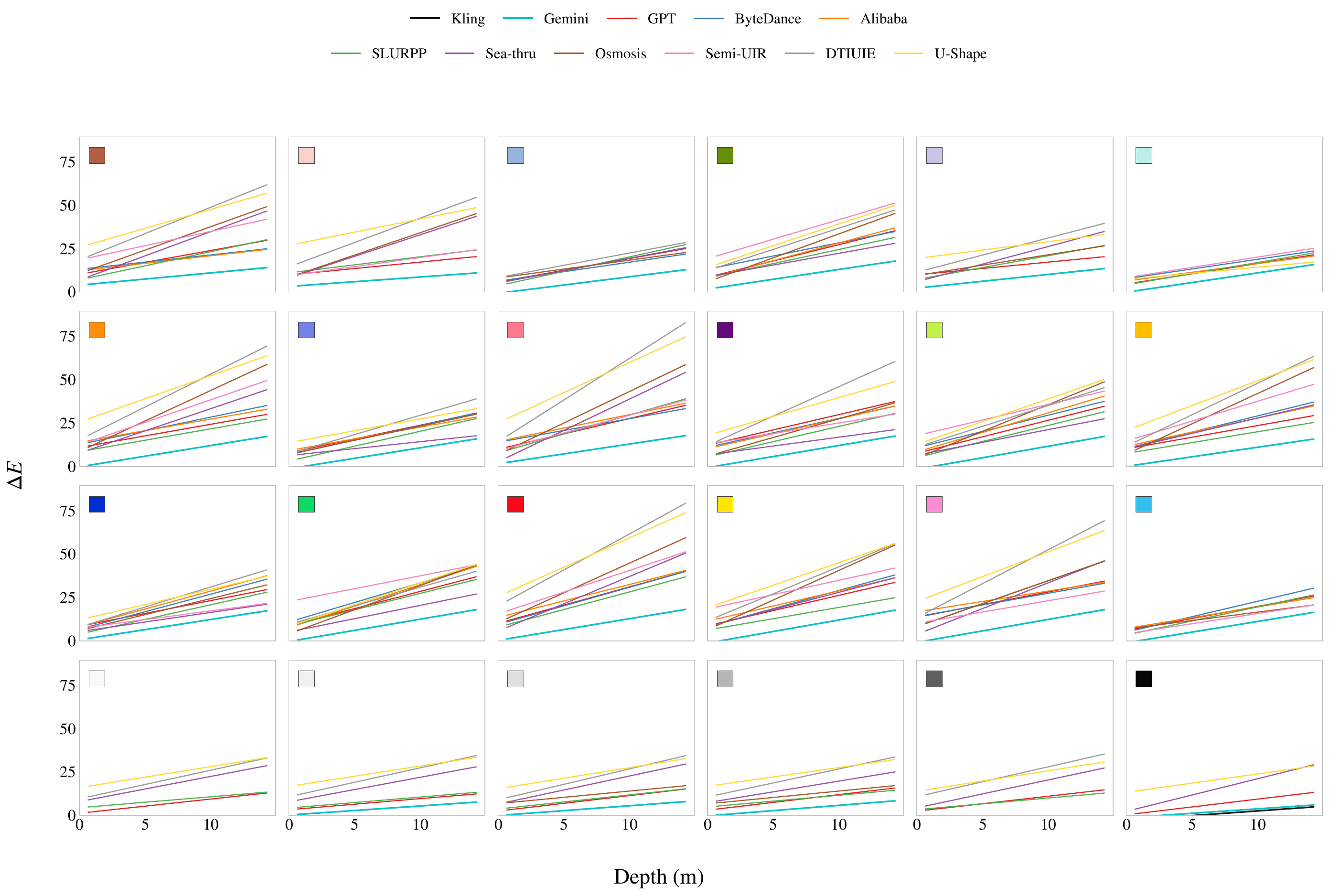}
    \caption{Model-predicted color-patch restoration error as a function of depth. Each panel corresponds to one ground-truth color patch, indicated by the colored square. Lines show method-specific predicted Delta E across the observed depth range under the reference water and dataset, and only method-patch depth slopes significant at \(p<0.001\) are plotted. The model includes method-by-patch-by-depth fixed effects, so slopes differ by method and color patch.}
    \label{fig:rq3_patch_depth_profile}
\end{figure*}

\textbf{Sample size for evaluation experiments.}
\begin{itemize}
    \item Tab.~\ref{tab:deltae-depth-bin-counts} shows the sample sizes for mean $\Delta E$ values per depth-bin heatmap fig.~\ref{fig:deltae-depth-heatmap} in the main paper. These samples sizes are the same for patch-level analysis (RQ3), Fig.~\ref{fig:rq3_patch_ranked_bars} for color patch performance in the main paper as all the patches are present in all the observations (method restorations).
    \item For RQs 1-2, the model fit on 4,150 observations (method restorations) from 626 unique image conditions (set, scene, water, frame). 
    \item Reference depth set at mean depth 3.5 m (observed range 0.68-14.26 m).  Fig~\ref{fig:depth_slope} shows the depth sensitivity curves per method. Fig~\ref{fig:histogram_depth_coverage} shows depth coverage histograms. These correspond to Tab.~\ref{tab:rq1_adjusted_accuracy} in the main paper.
    \item For RQ 4, Tab.~\ref{tab:rq3_water_sensitivity_cell_counts} shows the sample sizes (method restorations) for each water type, corresponding to the heatmap Fig.~\ref{fig:water_sensitivity} in the main paper.
    \item For RQ5, tab.~\ref{tab:rq4_own_dataset_bias_unique_image_cell_counts} shows the count of samples per dataset for each VLM restorer. The mean \(\Delta_{E}\) score changes reported in Fig.~\ref{fig:rq4_bias_heatmap} in the main paper are based on these sample sizes; for example, for Gemini, its own-family set Veo contains 118 samples; \(\Delta_{E}\) of this set versus mean \(\Delta_{E}\) of others (i.e., sum count of Kling, Sora, Qwen, and Seedance which is 380).
    \item Fig.~\ref{fig:rq3_patch_depth_profile} shows depth-slope curves per \textit{color patch} for all the methods.
\end{itemize}
\newpage

\begin{table}[]
\centering
\small
\setlength{\tabcolsep}{4pt}
\renewcommand{\arraystretch}{1.15}
\caption{Sample sizes for the RQ4 water-sensitivity heatmap cells. Counts are chart-level observations in each method-by-water cell, ordered to match the heatmap Fig.~\ref{fig:water_sensitivity} in the main paper.}
\label{tab:rq3_water_sensitivity_cell_counts}
\begin{tabular}{l|rrrrr}
\toprule
\rowcolor{gray!15}
\textbf{Method} & \textbf{I} & \textbf{II} & \textbf{III} & \textbf{IV} & \textbf{V} \\
\midrule
U-shape & 98 & 76 & 51 & 47 & 33 \\
Dtiuie & 62 & 25 & 19 & 18 & 13 \\
Semi-uir & 73 & 48 & 35 & 31 & 18 \\
Seathru & 103 & 71 & 51 & 45 & 38 \\
Osmosis & 110 & 86 & 72 & 61 & 44 \\
Qwen & 132 & 116 & 88 & 79 & 62 \\
Seedream & 143 & 117 & 92 & 83 & 65 \\
Gpt & 86 & 73 & 71 & 59 & 44 \\
Slurpp & 133 & 111 & 90 & 93 & 75 \\
Gemini & 136 & 117 & 95 & 86 & 64 \\
Kling & 135 & 122 & 102 & 93 & 60 \\
\rowcolor{orange!14}
\textbf{Total} & \textbf{1211} & \textbf{962} & \textbf{766} & \textbf{695} & \textbf{516} \\
\bottomrule
\end{tabular}
\end{table}

\begin{table}[]
\centering
\small
\setlength{\tabcolsep}{4pt}
\renewcommand{\arraystretch}{1.15}
\caption{Unique-image counts for each own-dataset bias heatmap cell (RQ5) in Fig.~\ref{fig:rq4_bias_heatmap} in the main paper. \colorbox{red!14}{Color} denotes count of own-family/generated set. The other-total column sums each row excluding the method's own generated set.}
\label{tab:rq4_own_dataset_bias_unique_image_cell_counts}
\begin{tabular}{l|rrrrr|r}
\toprule
\rowcolor{gray!15}
\textbf{Method} & \textbf{Veo} & \textbf{Kling} & \textbf{Sora} & \textbf{Qwen} & \textbf{Seedance} & \textbf{Other total} \\
\midrule
Gemini & \cellcolor{red!14}118 & 110 & 137 & 68 & 65 & 380 \\
Kling & 115 & \cellcolor{red!14}114 & 140 & 77 & 66 & 398 \\
Gpt & 87 & 76 & \cellcolor{red!14}82 & 58 & 30 & 251 \\
Qwen & 132 & 91 & 121 & \cellcolor{red!14}72 & 61 & 405 \\
Seedream & 123 & 103 & 142 & 71 & \cellcolor{red!14}61 & 439 \\
\rowcolor{orange!14}
\textbf{Total} & \textbf{575} & \textbf{494} & \textbf{622} & \textbf{346} & \textbf{283} & \textbf{1873} \\
\bottomrule
\end{tabular}
\end{table}

\clearpage

\clearpage
\section*{Supplementary Note 6}
\textbf{Reference conditions for LMEM.}
The LMEM estimates performance relative to a reference condition. We use the following reference levels: \textbf{restorer: Gemini}, \textbf{dataset: Veo-set}, and \textbf{water type: I}. The model first predicts the expected error for Gemini under these conditions, then estimates how performance changes with restoration method, dataset, or water type relative to this baseline.
The reference levels were chosen based on the following observations. 

\begin{figure}[]
    \centering
    \includegraphics[width=0.95\linewidth]{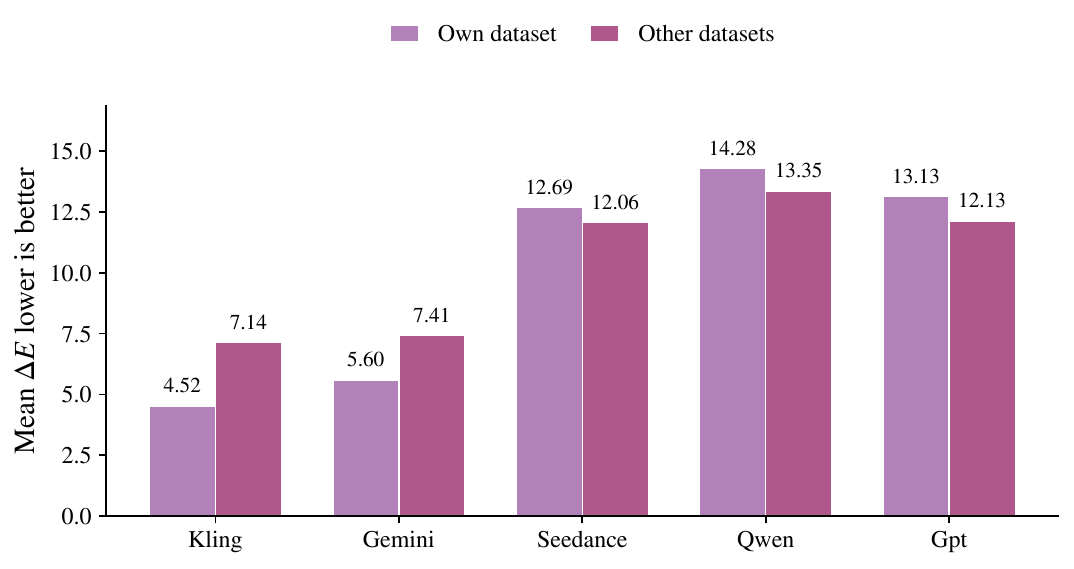}
    \caption{Gemini is used as the LMEM reference method because it combines low restoration error with lower own-dataset bias than Kling.}
    \label{fig:reference_gemini}
\end{figure}

Fig.~\ref{fig:reference_gemini} compares each method's mean (\(\Delta E\)) on its own dataset versus the other datasets. Kling achieves the lowest error but shows the strongest own-dataset bias. Gemini performs nearly as well while exhibiting less bias, making it a more suitable reference method.

\begin{figure}[]
    \centering
    \includegraphics[width=0.95\linewidth]{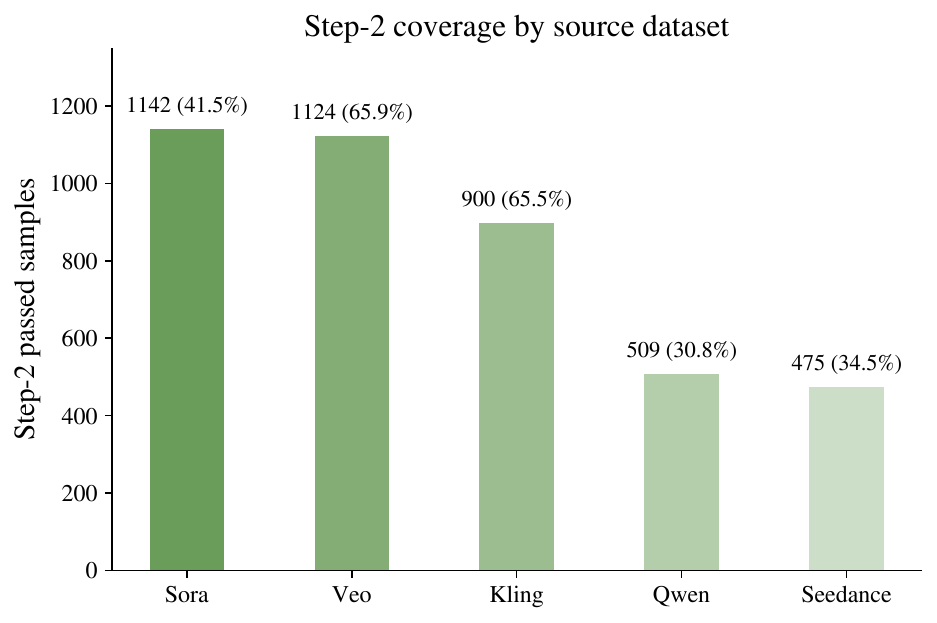}
    \caption{Veo-set is used as the LMEM reference dataset because it has the highest number of samples retained after the Step 1–2 evaluation filtering.}
    \label{fig:reference-veo}
\end{figure}

Fig.~\ref{fig:reference-veo} shows the number of images retained after the Step 1–2 evaluation filtering. Veo-set has the highest number of retained samples and is therefore used as the reference dataset.

\begin{figure}[]
    \centering
    \includegraphics[width=0.95\linewidth]{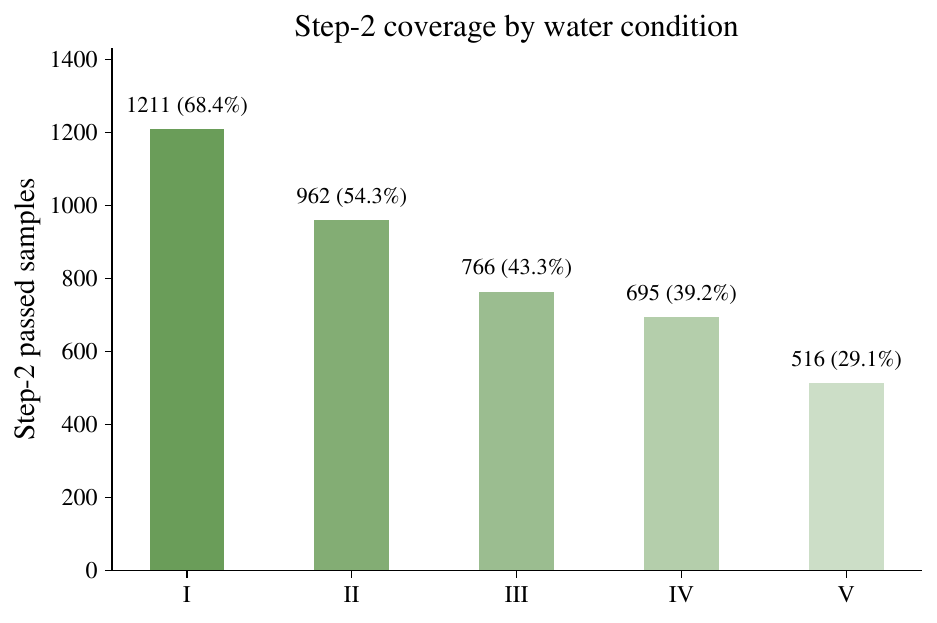}
    \caption{Water type I is used as the LMEM reference condition because it contains the highest number of samples retained after the Step 1–2 evaluation filtering.}
    \label{fig:reference-waterI}
\end{figure}

Fig.~\ref{fig:reference-waterI} shows the distribution of retained samples across water types. Water type I contains the largest number of observations and is therefore selected as the reference water condition.

\newpage

\clearpage
\section*{Supplementary Note 7}
Linear mixed-effects model (LMEM) results. Tab.~\ref{tab:supp_lmem_omnibus} reports the omnibus tests for the main effects and interactions discussed in the paper. Tab.~\ref{tab:supp_lmem_fixed_effects} reports the fixed-effect estimates, Tab.~\ref{tab:supp_lmem_random_stats} summarizes the random effects and model statistics.
\begin{table}[H]
\caption{Omnibus Wald tests for the LMEM fixed-effect terms corresponding to each research question.}
\label{tab:supp_lmem_omnibus}
\begin{tabular}{lllrrr}
\toprule
RQ & Model & Effect & $\chi^2$ & df & $p$ \\
\midrule
1 & Chart & Method & 1883.67 & 10 & $<0.001$ \\
2 & Chart & Method x depth & 306.70 & 10 & $<0.001$ \\
3 & Patch & Method x patch & 15937.30 & 230 & $<0.001$ \\
4 & Chart & Method x water & 144.63 & 40 & $<0.001$ \\
5 & Chart & Method x dataset & 566.85 & 40 & $<0.001$ \\
\bottomrule
\end{tabular}
\vspace{2pt}
\footnotesize
\noindent\emph{Note.} Each row reports an omnibus Wald test for a fixed effect. The Model column indicates whether the test is from the chart-level $\Delta E$ LMEM or the patch-level $\log(1+\Delta E)$ LMEM. df is the degrees of freedom (number of independent model constraints tested, usually the number of coefficients in that term), $\chi^2$ is the Wald statistic (larger values relative to df indicate stronger departure from the null hypothesis) and $p$ is the corresponding significance level. $p<0.001$ indicates strong evidence that the corresponding effect is present. Fixed-effect coefficient estimates are reported in Tab.~\ref{tab:supp_lmem_fixed_effects}.
\end{table}

\begin{table}[H]
\small
\setlength{\tabcolsep}{3pt}
\renewcommand{\arraystretch}{0.98}
\caption{Random-effect estimates and model statistics for the LMEM. Random-effect entries report variance estimates with standard errors in parentheses.}
\label{tab:supp_lmem_random_stats}
\begin{tabular}{@{}p{0.12\linewidth}p{0.27\linewidth}p{0.53\linewidth}@{}}
\toprule
Model & Statistic & Value \\
\midrule
Chart & Scale & Delta E \\
 & Observations & 4150 \\
 & Groups & 626 \\
 & Residual scale & 8.1317 \\
 & Log-likelihood & -10584.3104 \\
 & Converged & Yes \\
 & Random effects & Group Var=3.400 (SE=0.111) \\
\addlinespace[2pt]
Patch & Scale & log(1 + delta\_E); exported scores are back-transformed \\
 & Observations & 99600 \\
 & Groups & 626 \\
 & Residual scale & 0.2371 \\
 & Log-likelihood & -73018.1625 \\
 & Converged & Yes \\
 & Random effects & Group Var=0.055 (SE=0.007); patch Var=0.010 (SE=0.001) \\
\bottomrule
\end{tabular}
\end{table}

\begingroup
\scriptsize
\setlength{\tabcolsep}{2pt}
\renewcommand{\arraystretch}{0.95}
\onecolumn
\begin{longtable}{@{}p{0.14\linewidth}p{0.07\linewidth}p{0.28\linewidth}rrrr@{}}
\caption{Fixed-effect estimates for the LMEM terms corresponding to each research question. The chart model is fit on $\Delta E$; the patch model is fit on $\log(1+\Delta E)$. Method, water, dataset, and patch contrasts are relative to Gemini, water type I, Veo-set, and patch 0, respectively.}\label{tab:supp_lmem_fixed_effects}\\
\toprule
RQ/effect & Model & Contrast & Estimate & SE & $z$ & $p$ \\
\midrule
\endfirsthead
\toprule
RQ/effect & Model & Contrast & Estimate & SE & $z$ & $p$ \\
\midrule
\endhead
1 Method & Chart & Qwen vs Gemini & 6.499 & 0.481 & 13.513 & $<0.001$ \\
 & Chart & Seedream vs Gemini & 6.085 & 0.480 & 12.672 & $<0.001$ \\
 & Chart & Dtiuie vs Gemini & 15.845 & 0.651 & 24.339 & $<0.001$ \\
 & Chart & Gpt vs Gemini & 6.266 & 0.546 & 11.480 & $<0.001$ \\
 & Chart & Kling vs Gemini & 0.748 & 0.488 & 1.532 & 0.125 \\
 & Chart & Osmosis vs Gemini & 8.664 & 0.511 & 16.970 & $<0.001$ \\
 & Chart & Seathru vs Gemini & 9.220 & 0.512 & 17.990 & $<0.001$ \\
 & Chart & Semi-uir vs Gemini & 13.272 & 0.593 & 22.391 & $<0.001$ \\
 & Chart & Slurpp vs Gemini & 4.964 & 0.482 & 10.307 & $<0.001$ \\
 & Chart & U-shape vs Gemini & 17.744 & 0.537 & 33.022 & $<0.001$ \\
2 Method x depth & Chart & Gemini depth slope & 0.497 & 0.075 & 6.648 & $<0.001$ \\
 & Chart & Qwen slope difference vs Gemini & -0.020 & 0.099 & -0.200 & 0.841 \\
 & Chart & Seedream slope difference vs Gemini & 0.079 & 0.091 & 0.864 & 0.388 \\
 & Chart & Dtiuie slope difference vs Gemini & 1.908 & 0.189 & 10.075 & $<0.001$ \\
 & Chart & Gpt slope difference vs Gemini & 0.179 & 0.105 & 1.704 & 0.088 \\
 & Chart & Kling slope difference vs Gemini & -0.259 & 0.090 & -2.893 & 0.004 \\
 & Chart & Osmosis slope difference vs Gemini & 0.762 & 0.110 & 6.899 & $<0.001$ \\
 & Chart & Seathru slope difference vs Gemini & 0.869 & 0.121 & 7.189 & $<0.001$ \\
 & Chart & Semi-uir slope difference vs Gemini & -0.012 & 0.149 & -0.078 & 0.938 \\
 & Chart & Slurpp slope difference vs Gemini & 0.253 & 0.089 & 2.834 & 0.005 \\
 & Chart & U-shape slope difference vs Gemini & 0.977 & 0.117 & 8.360 & $<0.001$ \\
4 Method x water & Chart & Qwen x V & 0.183 & 0.634 & 0.288 & 0.773 \\
 & Chart & Seedream x V & -0.162 & 0.625 & -0.260 & 0.795 \\
 & Chart & Dtiuie x V & 1.492 & 1.069 & 1.396 & 0.163 \\
 & Chart & Gpt x V & -0.178 & 0.704 & -0.253 & 0.800 \\
 & Chart & Kling x V & -1.298 & 0.638 & -2.035 & 0.042 \\
 & Chart & Osmosis x V & 0.676 & 0.696 & 0.971 & 0.331 \\
 & Chart & Seathru x V & -0.088 & 0.729 & -0.120 & 0.904 \\
 & Chart & Semi-uir x V & 0.228 & 0.918 & 0.248 & 0.804 \\
 & Chart & Slurpp x V & 0.992 & 0.615 & 1.615 & 0.106 \\
 & Chart & U-shape x V & -1.971 & 0.752 & -2.622 & 0.009 \\
 & Chart & Qwen x III & 2.005 & 0.558 & 3.592 & $<0.001$ \\
 & Chart & Seedream x III & -0.075 & 0.550 & -0.136 & 0.892 \\
 & Chart & Dtiuie x III & -0.536 & 0.905 & -0.593 & 0.554 \\
 & Chart & Gpt x III & -0.479 & 0.607 & -0.790 & 0.429 \\
 & Chart & Kling x III & -0.961 & 0.543 & -1.770 & 0.077 \\
 & Chart & Osmosis x III & 1.630 & 0.597 & 2.732 & 0.006 \\
 & Chart & Seathru x III & 0.123 & 0.644 & 0.191 & 0.848 \\
 & Chart & Semi-uir x III & -3.124 & 0.737 & -4.238 & $<0.001$ \\
 & Chart & Slurpp x III & 0.382 & 0.555 & 0.689 & 0.491 \\
 & Chart & U-shape x III & 0.860 & 0.642 & 1.339 & 0.180 \\
 & Chart & Qwen x II & 0.737 & 0.515 & 1.431 & 0.152 \\
 & Chart & Seedream x II & 0.320 & 0.511 & 0.627 & 0.531 \\
 & Chart & Dtiuie x II & 0.300 & 0.810 & 0.371 & 0.711 \\
 & Chart & Gpt x II & 0.451 & 0.589 & 0.766 & 0.444 \\
 & Chart & Kling x II & -0.305 & 0.512 & -0.596 & 0.551 \\
 & Chart & Osmosis x II & 0.905 & 0.556 & 1.627 & 0.104 \\
 & Chart & Seathru x II & 0.143 & 0.583 & 0.245 & 0.806 \\
 & Chart & Semi-uir x II & -1.354 & 0.662 & -2.044 & 0.041 \\
 & Chart & Slurpp x II & 0.112 & 0.520 & 0.215 & 0.829 \\
 & Chart & U-shape x II & -0.499 & 0.577 & -0.864 & 0.387 \\
 & Chart & Qwen x IV & 0.590 & 0.578 & 1.021 & 0.307 \\
 & Chart & Seedream x IV & -0.514 & 0.569 & -0.904 & 0.366 \\
 & Chart & Dtiuie x IV & 0.358 & 0.921 & 0.388 & 0.698 \\
 & Chart & Gpt x IV & -1.277 & 0.641 & -1.991 & 0.047 \\
 & Chart & Kling x IV & -1.624 & 0.561 & -2.896 & 0.004 \\
 & Chart & Osmosis x IV & 1.173 & 0.626 & 1.875 & 0.061 \\
 & Chart & Seathru x IV & -0.101 & 0.671 & -0.150 & 0.881 \\
 & Chart & Semi-uir x IV & -3.547 & 0.768 & -4.618 & $<0.001$ \\
 & Chart & Slurpp x IV & 0.061 & 0.562 & 0.109 & 0.913 \\
 & Chart & U-shape x IV & -0.087 & 0.669 & -0.130 & 0.896 \\
3 Method x patch & Patch & Qwen x patch 1 & 0.582 & 0.044 & 13.168 & $<0.001$ \\
 & Patch & Seedream x patch 1 & 0.457 & 0.044 & 10.460 & $<0.001$ \\
 & Patch & Dtiuie x patch 1 & 0.278 & 0.067 & 4.143 & $<0.001$ \\
 & Patch & Gpt x patch 1 & 0.295 & 0.049 & 6.027 & $<0.001$ \\
 & Patch & Kling x patch 1 & 0.356 & 0.043 & 8.197 & $<0.001$ \\
 & Patch & Osmosis x patch 1 & 0.326 & 0.047 & 6.895 & $<0.001$ \\
 & Patch & Seathru x patch 1 & 0.333 & 0.050 & 6.638 & $<0.001$ \\
 & Patch & Semi-uir x patch 1 & 0.129 & 0.058 & 2.245 & 0.025 \\
 & Patch & Slurpp x patch 1 & 0.422 & 0.044 & 9.666 & $<0.001$ \\
 & Patch & U-shape x patch 1 & 0.377 & 0.050 & 7.497 & $<0.001$ \\
 & Patch & Qwen x patch 10 & 0.546 & 0.044 & 12.357 & $<0.001$ \\
 & Patch & Seedream x patch 10 & 0.236 & 0.044 & 5.407 & $<0.001$ \\
 & Patch & Dtiuie x patch 10 & 0.417 & 0.067 & 6.223 & $<0.001$ \\
 & Patch & Gpt x patch 10 & 0.248 & 0.049 & 5.062 & $<0.001$ \\
 & Patch & Kling x patch 10 & 0.053 & 0.043 & 1.224 & 0.221 \\
 & Patch & Osmosis x patch 10 & 0.322 & 0.047 & 6.810 & $<0.001$ \\
 & Patch & Seathru x patch 10 & 0.186 & 0.050 & 3.704 & $<0.001$ \\
 & Patch & Semi-uir x patch 10 & 0.168 & 0.058 & 2.926 & 0.003 \\
 & Patch & Slurpp x patch 10 & 0.431 & 0.044 & 9.887 & $<0.001$ \\
 & Patch & U-shape x patch 10 & 0.340 & 0.050 & 6.768 & $<0.001$ \\
 & Patch & Qwen x patch 11 & -0.071 & 0.044 & -1.604 & 0.109 \\
 & Patch & Seedream x patch 11 & -0.435 & 0.044 & -9.954 & $<0.001$ \\
 & Patch & Dtiuie x patch 11 & 0.010 & 0.067 & 0.145 & 0.885 \\
 & Patch & Gpt x patch 11 & -0.183 & 0.049 & -3.748 & $<0.001$ \\
 & Patch & Kling x patch 11 & 0.324 & 0.043 & 7.457 & $<0.001$ \\
 & Patch & Osmosis x patch 11 & 0.110 & 0.047 & 2.319 & 0.020 \\
 & Patch & Seathru x patch 11 & 0.397 & 0.050 & 7.920 & $<0.001$ \\
 & Patch & Semi-uir x patch 11 & -0.128 & 0.058 & -2.232 & 0.026 \\
 & Patch & Slurpp x patch 11 & 0.236 & 0.044 & 5.412 & $<0.001$ \\
 & Patch & U-shape x patch 11 & 0.162 & 0.050 & 3.214 & 0.001 \\
 & Patch & Qwen x patch 12 & 0.165 & 0.044 & 3.742 & $<0.001$ \\
 & Patch & Seedream x patch 12 & 0.223 & 0.044 & 5.114 & $<0.001$ \\
 & Patch & Dtiuie x patch 12 & -0.158 & 0.067 & -2.354 & 0.019 \\
 & Patch & Gpt x patch 12 & 0.012 & 0.049 & 0.243 & 0.808 \\
 & Patch & Kling x patch 12 & -0.094 & 0.043 & -2.155 & 0.031 \\
 & Patch & Osmosis x patch 12 & -0.152 & 0.047 & -3.214 & 0.001 \\
 & Patch & Seathru x patch 12 & -0.078 & 0.050 & -1.562 & 0.118 \\
 & Patch & Semi-uir x patch 12 & 0.215 & 0.058 & 3.735 & $<0.001$ \\
 & Patch & Slurpp x patch 12 & 0.204 & 0.044 & 4.667 & $<0.001$ \\
 & Patch & U-shape x patch 12 & -0.268 & 0.050 & -5.335 & $<0.001$ \\
 & Patch & Qwen x patch 13 & 0.574 & 0.044 & 12.987 & $<0.001$ \\
 & Patch & Seedream x patch 13 & 0.449 & 0.044 & 10.284 & $<0.001$ \\
 & Patch & Dtiuie x patch 13 & 0.089 & 0.067 & 1.332 & 0.183 \\
 & Patch & Gpt x patch 13 & 0.570 & 0.049 & 11.650 & $<0.001$ \\
 & Patch & Kling x patch 13 & 0.239 & 0.043 & 5.491 & $<0.001$ \\
 & Patch & Osmosis x patch 13 & 0.019 & 0.047 & 0.410 & 0.682 \\
 & Patch & Seathru x patch 13 & -0.039 & 0.050 & -0.781 & 0.435 \\
 & Patch & Semi-uir x patch 13 & 0.173 & 0.058 & 3.011 & 0.003 \\
 & Patch & Slurpp x patch 13 & 0.451 & 0.044 & 10.324 & $<0.001$ \\
 & Patch & U-shape x patch 13 & 0.136 & 0.050 & 2.701 & 0.007 \\
 & Patch & Qwen x patch 14 & 0.687 & 0.044 & 15.545 & $<0.001$ \\
 & Patch & Seedream x patch 14 & 0.401 & 0.044 & 9.189 & $<0.001$ \\
 & Patch & Dtiuie x patch 14 & 0.233 & 0.067 & 3.476 & 0.001 \\
 & Patch & Gpt x patch 14 & 0.307 & 0.049 & 6.269 & $<0.001$ \\
 & Patch & Kling x patch 14 & 0.564 & 0.043 & 12.978 & $<0.001$ \\
 & Patch & Osmosis x patch 14 & 0.374 & 0.047 & 7.898 & $<0.001$ \\
 & Patch & Seathru x patch 14 & 0.428 & 0.050 & 8.533 & $<0.001$ \\
 & Patch & Semi-uir x patch 14 & 0.537 & 0.058 & 9.334 & $<0.001$ \\
 & Patch & Slurpp x patch 14 & 0.348 & 0.044 & 7.968 & $<0.001$ \\
 & Patch & U-shape x patch 14 & 0.323 & 0.050 & 6.420 & $<0.001$ \\
 & Patch & Qwen x patch 15 & 0.108 & 0.044 & 2.443 & 0.015 \\
 & Patch & Seedream x patch 15 & -0.288 & 0.044 & -6.595 & $<0.001$ \\
 & Patch & Dtiuie x patch 15 & 0.094 & 0.067 & 1.408 & 0.159 \\
 & Patch & Gpt x patch 15 & -0.126 & 0.049 & -2.573 & 0.010 \\
 & Patch & Kling x patch 15 & 0.362 & 0.043 & 8.346 & $<0.001$ \\
 & Patch & Osmosis x patch 15 & 0.090 & 0.047 & 1.900 & 0.057 \\
 & Patch & Seathru x patch 15 & 0.360 & 0.050 & 7.171 & $<0.001$ \\
 & Patch & Semi-uir x patch 15 & -0.102 & 0.058 & -1.773 & 0.076 \\
 & Patch & Slurpp x patch 15 & 0.305 & 0.044 & 6.981 & $<0.001$ \\
 & Patch & U-shape x patch 15 & 0.202 & 0.050 & 4.022 & $<0.001$ \\
 & Patch & Qwen x patch 16 & 0.182 & 0.044 & 4.117 & $<0.001$ \\
 & Patch & Seedream x patch 16 & -0.062 & 0.044 & -1.424 & 0.154 \\
 & Patch & Dtiuie x patch 16 & -0.261 & 0.067 & -3.901 & $<0.001$ \\
 & Patch & Gpt x patch 16 & 0.035 & 0.049 & 0.709 & 0.478 \\
 & Patch & Kling x patch 16 & 0.097 & 0.043 & 2.226 & 0.026 \\
 & Patch & Osmosis x patch 16 & -0.084 & 0.047 & -1.764 & 0.078 \\
 & Patch & Seathru x patch 16 & -0.051 & 0.050 & -1.010 & 0.312 \\
 & Patch & Semi-uir x patch 16 & -0.577 & 0.058 & -10.040 & $<0.001$ \\
 & Patch & Slurpp x patch 16 & 0.259 & 0.044 & 5.946 & $<0.001$ \\
 & Patch & U-shape x patch 16 & -0.130 & 0.050 & -2.580 & 0.010 \\
 & Patch & Qwen x patch 17 & 0.714 & 0.044 & 16.164 & $<0.001$ \\
 & Patch & Seedream x patch 17 & 0.635 & 0.044 & 14.530 & $<0.001$ \\
 & Patch & Dtiuie x patch 17 & 0.266 & 0.067 & 3.973 & $<0.001$ \\
 & Patch & Gpt x patch 17 & 0.395 & 0.049 & 8.070 & $<0.001$ \\
 & Patch & Kling x patch 17 & 0.381 & 0.043 & 8.780 & $<0.001$ \\
 & Patch & Osmosis x patch 17 & 0.325 & 0.047 & 6.860 & $<0.001$ \\
 & Patch & Seathru x patch 17 & 0.281 & 0.050 & 5.601 & $<0.001$ \\
 & Patch & Semi-uir x patch 17 & 0.603 & 0.058 & 10.488 & $<0.001$ \\
 & Patch & Slurpp x patch 17 & 0.516 & 0.044 & 11.832 & $<0.001$ \\
 & Patch & U-shape x patch 17 & 0.152 & 0.050 & 3.017 & 0.003 \\
 & Patch & Qwen x patch 18 & 0.839 & 0.044 & 18.980 & $<0.001$ \\
 & Patch & Seedream x patch 18 & 0.597 & 0.044 & 13.679 & $<0.001$ \\
 & Patch & Dtiuie x patch 18 & 0.296 & 0.067 & 4.419 & $<0.001$ \\
 & Patch & Gpt x patch 18 & 0.551 & 0.049 & 11.258 & $<0.001$ \\
 & Patch & Kling x patch 18 & 0.497 & 0.043 & 11.436 & $<0.001$ \\
 & Patch & Osmosis x patch 18 & 0.291 & 0.047 & 6.146 & $<0.001$ \\
 & Patch & Seathru x patch 18 & 0.188 & 0.050 & 3.748 & $<0.001$ \\
 & Patch & Semi-uir x patch 18 & -0.038 & 0.058 & -0.669 & 0.504 \\
 & Patch & Slurpp x patch 18 & 0.763 & 0.044 & 17.489 & $<0.001$ \\
 & Patch & U-shape x patch 18 & 0.427 & 0.050 & 8.499 & $<0.001$ \\
 & Patch & Qwen x patch 19 & 0.131 & 0.044 & 2.953 & 0.003 \\
 & Patch & Seedream x patch 19 & -0.330 & 0.044 & -7.552 & $<0.001$ \\
 & Patch & Dtiuie x patch 19 & 0.216 & 0.067 & 3.216 & 0.001 \\
 & Patch & Gpt x patch 19 & -0.156 & 0.049 & -3.192 & 0.001 \\
 & Patch & Kling x patch 19 & 0.387 & 0.043 & 8.913 & $<0.001$ \\
 & Patch & Osmosis x patch 19 & 0.088 & 0.047 & 1.861 & 0.063 \\
 & Patch & Seathru x patch 19 & 0.262 & 0.050 & 5.226 & $<0.001$ \\
 & Patch & Semi-uir x patch 19 & 0.178 & 0.058 & 3.100 & 0.002 \\
 & Patch & Slurpp x patch 19 & 0.198 & 0.044 & 4.546 & $<0.001$ \\
 & Patch & U-shape x patch 19 & 0.143 & 0.050 & 2.843 & 0.004 \\
 & Patch & Qwen x patch 2 & 0.288 & 0.044 & 6.527 & $<0.001$ \\
 & Patch & Seedream x patch 2 & 0.083 & 0.044 & 1.899 & 0.058 \\
 & Patch & Dtiuie x patch 2 & -0.367 & 0.067 & -5.482 & $<0.001$ \\
 & Patch & Gpt x patch 2 & 0.067 & 0.049 & 1.375 & 0.169 \\
 & Patch & Kling x patch 2 & 0.037 & 0.043 & 0.853 & 0.393 \\
 & Patch & Osmosis x patch 2 & -0.228 & 0.047 & -4.812 & $<0.001$ \\
 & Patch & Seathru x patch 2 & -0.278 & 0.050 & -5.541 & $<0.001$ \\
 & Patch & Semi-uir x patch 2 & -0.478 & 0.058 & -8.307 & $<0.001$ \\
 & Patch & Slurpp x patch 2 & 0.158 & 0.044 & 3.620 & $<0.001$ \\
 & Patch & U-shape x patch 2 & -0.359 & 0.050 & -7.133 & $<0.001$ \\
 & Patch & Qwen x patch 20 & 0.034 & 0.044 & 0.777 & 0.437 \\
 & Patch & Seedream x patch 20 & 0.029 & 0.044 & 0.674 & 0.501 \\
 & Patch & Dtiuie x patch 20 & -0.707 & 0.067 & -10.548 & $<0.001$ \\
 & Patch & Gpt x patch 20 & -0.098 & 0.049 & -2.009 & 0.045 \\
 & Patch & Kling x patch 20 & 0.482 & 0.043 & 11.094 & $<0.001$ \\
 & Patch & Osmosis x patch 20 & -0.550 & 0.047 & -11.613 & $<0.001$ \\
 & Patch & Seathru x patch 20 & -0.185 & 0.050 & -3.686 & $<0.001$ \\
 & Patch & Semi-uir x patch 20 & -0.329 & 0.058 & -5.723 & $<0.001$ \\
 & Patch & Slurpp x patch 20 & 0.160 & 0.044 & 3.677 & $<0.001$ \\
 & Patch & U-shape x patch 20 & -0.758 & 0.050 & -15.070 & $<0.001$ \\
 & Patch & Qwen x patch 21 & 0.580 & 0.044 & 13.115 & $<0.001$ \\
 & Patch & Seedream x patch 21 & 0.419 & 0.044 & 9.601 & $<0.001$ \\
 & Patch & Dtiuie x patch 21 & 0.171 & 0.067 & 2.554 & 0.011 \\
 & Patch & Gpt x patch 21 & 0.277 & 0.049 & 5.657 & $<0.001$ \\
 & Patch & Kling x patch 21 & 0.499 & 0.043 & 11.477 & $<0.001$ \\
 & Patch & Osmosis x patch 21 & 0.332 & 0.047 & 7.009 & $<0.001$ \\
 & Patch & Seathru x patch 21 & 0.422 & 0.050 & 8.423 & $<0.001$ \\
 & Patch & Semi-uir x patch 21 & 0.307 & 0.058 & 5.332 & $<0.001$ \\
 & Patch & Slurpp x patch 21 & 0.378 & 0.044 & 8.659 & $<0.001$ \\
 & Patch & U-shape x patch 21 & 0.314 & 0.050 & 6.240 & $<0.001$ \\
 & Patch & Qwen x patch 22 & 0.287 & 0.044 & 6.484 & $<0.001$ \\
 & Patch & Seedream x patch 22 & 0.103 & 0.044 & 2.358 & 0.018 \\
 & Patch & Dtiuie x patch 22 & -0.635 & 0.067 & -9.477 & $<0.001$ \\
 & Patch & Gpt x patch 22 & 0.126 & 0.049 & 2.565 & 0.010 \\
 & Patch & Kling x patch 22 & 0.358 & 0.043 & 8.244 & $<0.001$ \\
 & Patch & Osmosis x patch 22 & -0.040 & 0.047 & -0.838 & 0.402 \\
 & Patch & Seathru x patch 22 & 0.031 & 0.050 & 0.628 & 0.530 \\
 & Patch & Semi-uir x patch 22 & -0.542 & 0.058 & -9.429 & $<0.001$ \\
 & Patch & Slurpp x patch 22 & 0.278 & 0.044 & 6.373 & $<0.001$ \\
 & Patch & U-shape x patch 22 & -0.665 & 0.050 & -13.229 & $<0.001$ \\
 & Patch & Qwen x patch 23 & 0.046 & 0.044 & 1.041 & 0.298 \\
 & Patch & Seedream x patch 23 & -0.513 & 0.044 & -11.740 & $<0.001$ \\
 & Patch & Dtiuie x patch 23 & 0.676 & 0.067 & 10.083 & $<0.001$ \\
 & Patch & Gpt x patch 23 & -0.187 & 0.049 & -3.820 & $<0.001$ \\
 & Patch & Kling x patch 23 & 0.277 & 0.043 & 6.388 & $<0.001$ \\
 & Patch & Osmosis x patch 23 & 0.458 & 0.047 & 9.685 & $<0.001$ \\
 & Patch & Seathru x patch 23 & 0.217 & 0.050 & 4.321 & $<0.001$ \\
 & Patch & Semi-uir x patch 23 & 0.710 & 0.058 & 12.351 & $<0.001$ \\
 & Patch & Slurpp x patch 23 & 0.312 & 0.044 & 7.139 & $<0.001$ \\
 & Patch & U-shape x patch 23 & 0.393 & 0.050 & 7.824 & $<0.001$ \\
 & Patch & Qwen x patch 3 & -0.001 & 0.044 & -0.025 & 0.980 \\
 & Patch & Seedream x patch 3 & -0.525 & 0.044 & -12.011 & $<0.001$ \\
 & Patch & Dtiuie x patch 3 & 0.124 & 0.067 & 1.850 & 0.064 \\
 & Patch & Gpt x patch 3 & -0.327 & 0.049 & -6.675 & $<0.001$ \\
 & Patch & Kling x patch 3 & 0.366 & 0.043 & 8.433 & $<0.001$ \\
 & Patch & Osmosis x patch 3 & 0.236 & 0.047 & 4.981 & $<0.001$ \\
 & Patch & Seathru x patch 3 & 0.533 & 0.050 & 10.629 & $<0.001$ \\
 & Patch & Semi-uir x patch 3 & 0.115 & 0.058 & 1.996 & 0.046 \\
 & Patch & Slurpp x patch 3 & 0.336 & 0.044 & 7.709 & $<0.001$ \\
 & Patch & U-shape x patch 3 & 0.288 & 0.050 & 5.721 & $<0.001$ \\
 & Patch & Qwen x patch 4 & 0.170 & 0.044 & 3.848 & $<0.001$ \\
 & Patch & Seedream x patch 4 & -0.121 & 0.044 & -2.776 & 0.006 \\
 & Patch & Dtiuie x patch 4 & -0.190 & 0.067 & -2.839 & 0.005 \\
 & Patch & Gpt x patch 4 & -0.042 & 0.049 & -0.858 & 0.391 \\
 & Patch & Kling x patch 4 & 0.120 & 0.043 & 2.761 & 0.006 \\
 & Patch & Osmosis x patch 4 & -0.006 & 0.047 & -0.136 & 0.892 \\
 & Patch & Seathru x patch 4 & 0.085 & 0.050 & 1.702 & 0.089 \\
 & Patch & Semi-uir x patch 4 & -0.493 & 0.058 & -8.573 & $<0.001$ \\
 & Patch & Slurpp x patch 4 & 0.331 & 0.044 & 7.592 & $<0.001$ \\
 & Patch & U-shape x patch 4 & 0.091 & 0.050 & 1.817 & 0.069 \\
 & Patch & Qwen x patch 5 & 0.583 & 0.044 & 13.195 & $<0.001$ \\
 & Patch & Seedream x patch 5 & 0.371 & 0.044 & 8.489 & $<0.001$ \\
 & Patch & Dtiuie x patch 5 & -0.077 & 0.067 & -1.155 & 0.248 \\
 & Patch & Gpt x patch 5 & 0.351 & 0.049 & 7.166 & $<0.001$ \\
 & Patch & Kling x patch 5 & 0.483 & 0.043 & 11.113 & $<0.001$ \\
 & Patch & Osmosis x patch 5 & 0.164 & 0.047 & 3.473 & 0.001 \\
 & Patch & Seathru x patch 5 & 0.020 & 0.050 & 0.394 & 0.693 \\
 & Patch & Semi-uir x patch 5 & -0.331 & 0.058 & -5.748 & $<0.001$ \\
 & Patch & Slurpp x patch 5 & 0.392 & 0.044 & 8.974 & $<0.001$ \\
 & Patch & U-shape x patch 5 & 0.028 & 0.050 & 0.564 & 0.573 \\
 & Patch & Qwen x patch 6 & 0.508 & 0.044 & 11.492 & $<0.001$ \\
 & Patch & Seedream x patch 6 & 0.506 & 0.044 & 11.577 & $<0.001$ \\
 & Patch & Dtiuie x patch 6 & -0.055 & 0.067 & -0.820 & 0.412 \\
 & Patch & Gpt x patch 6 & 0.299 & 0.049 & 6.116 & $<0.001$ \\
 & Patch & Kling x patch 6 & 0.413 & 0.043 & 9.517 & $<0.001$ \\
 & Patch & Osmosis x patch 6 & 0.023 & 0.047 & 0.477 & 0.633 \\
 & Patch & Seathru x patch 6 & -0.057 & 0.050 & -1.146 & 0.252 \\
 & Patch & Semi-uir x patch 6 & 0.576 & 0.058 & 10.017 & $<0.001$ \\
 & Patch & Slurpp x patch 6 & 0.540 & 0.044 & 12.372 & $<0.001$ \\
 & Patch & U-shape x patch 6 & -0.304 & 0.050 & -6.042 & $<0.001$ \\
 & Patch & Qwen x patch 7 & -0.085 & 0.044 & -1.912 & 0.056 \\
 & Patch & Seedream x patch 7 & -0.543 & 0.044 & -12.428 & $<0.001$ \\
 & Patch & Dtiuie x patch 7 & 0.054 & 0.067 & 0.800 & 0.424 \\
 & Patch & Gpt x patch 7 & -0.258 & 0.049 & -5.262 & $<0.001$ \\
 & Patch & Kling x patch 7 & 0.323 & 0.043 & 7.430 & $<0.001$ \\
 & Patch & Osmosis x patch 7 & 0.124 & 0.047 & 2.610 & 0.009 \\
 & Patch & Seathru x patch 7 & 0.396 & 0.050 & 7.905 & $<0.001$ \\
 & Patch & Semi-uir x patch 7 & -0.069 & 0.058 & -1.204 & 0.229 \\
 & Patch & Slurpp x patch 7 & 0.198 & 0.044 & 4.537 & $<0.001$ \\
 & Patch & U-shape x patch 7 & 0.179 & 0.050 & 3.558 & $<0.001$ \\
 & Patch & Qwen x patch 8 & 0.435 & 0.044 & 9.834 & $<0.001$ \\
 & Patch & Seedream x patch 8 & 0.168 & 0.044 & 3.840 & $<0.001$ \\
 & Patch & Dtiuie x patch 8 & -0.144 & 0.067 & -2.148 & 0.032 \\
 & Patch & Gpt x patch 8 & 0.244 & 0.049 & 4.991 & $<0.001$ \\
 & Patch & Kling x patch 8 & 0.470 & 0.043 & 10.811 & $<0.001$ \\
 & Patch & Osmosis x patch 8 & 0.160 & 0.047 & 3.382 & 0.001 \\
 & Patch & Seathru x patch 8 & 0.108 & 0.050 & 2.152 & 0.031 \\
 & Patch & Semi-uir x patch 8 & -0.494 & 0.058 & -8.584 & $<0.001$ \\
 & Patch & Slurpp x patch 8 & 0.411 & 0.044 & 9.427 & $<0.001$ \\
 & Patch & U-shape x patch 8 & -0.030 & 0.050 & -0.591 & 0.555 \\
 & Patch & Qwen x patch 9 & 0.440 & 0.044 & 9.950 & $<0.001$ \\
 & Patch & Seedream x patch 9 & 0.288 & 0.044 & 6.584 & $<0.001$ \\
 & Patch & Dtiuie x patch 9 & 0.039 & 0.067 & 0.587 & 0.557 \\
 & Patch & Gpt x patch 9 & 0.120 & 0.049 & 2.449 & 0.014 \\
 & Patch & Kling x patch 9 & 0.126 & 0.043 & 2.912 & 0.004 \\
 & Patch & Osmosis x patch 9 & 0.041 & 0.047 & 0.868 & 0.385 \\
 & Patch & Seathru x patch 9 & -0.043 & 0.050 & -0.851 & 0.395 \\
 & Patch & Semi-uir x patch 9 & -0.317 & 0.058 & -5.508 & $<0.001$ \\
 & Patch & Slurpp x patch 9 & 0.351 & 0.044 & 8.043 & $<0.001$ \\
 & Patch & U-shape x patch 9 & 0.204 & 0.050 & 4.048 & $<0.001$ \\
5 Method x dataset & Chart & Qwen x Kling & -0.311 & 0.591 & -0.526 & 0.599 \\
 & Chart & Seedream x Kling & -1.688 & 0.585 & -2.885 & 0.004 \\
 & Chart & Dtiuie x Kling & -6.913 & 0.850 & -8.134 & $<0.001$ \\
 & Chart & Gpt x Kling & -0.274 & 0.652 & -0.421 & 0.674 \\
 & Chart & Kling x Kling & -1.595 & 0.580 & -2.752 & 0.006 \\
 & Chart & Osmosis x Kling & -1.115 & 0.637 & -1.751 & 0.080 \\
 & Chart & Seathru x Kling & -0.701 & 0.636 & -1.102 & 0.271 \\
 & Chart & Semi-uir x Kling & -1.573 & 0.708 & -2.221 & 0.026 \\
 & Chart & Slurpp x Kling & -1.249 & 0.571 & -2.188 & 0.029 \\
 & Chart & U-shape x Kling & -3.734 & 0.655 & -5.698 & $<0.001$ \\
 & Chart & Qwen x Qwen & 0.605 & 0.612 & 0.988 & 0.323 \\
 & Chart & Seedream x Qwen & -0.801 & 0.617 & -1.298 & 0.194 \\
 & Chart & Dtiuie x Qwen & 0.275 & 1.181 & 0.233 & 0.816 \\
 & Chart & Gpt x Qwen & 0.871 & 0.665 & 1.310 & 0.190 \\
 & Chart & Kling x Qwen & 0.206 & 0.619 & 0.334 & 0.739 \\
 & Chart & Osmosis x Qwen & -0.326 & 0.797 & -0.409 & 0.682 \\
 & Chart & Seathru x Qwen & -1.025 & 0.860 & -1.192 & 0.233 \\
 & Chart & Semi-uir x Qwen & 0.558 & 1.055 & 0.529 & 0.597 \\
 & Chart & Slurpp x Qwen & 0.328 & 0.628 & 0.522 & 0.602 \\
 & Chart & U-shape x Qwen & 2.037 & 0.734 & 2.774 & 0.006 \\
 & Chart & Qwen x Seedance & -1.648 & 0.633 & -2.603 & 0.009 \\
 & Chart & Seedream x Seedance & -1.024 & 0.635 & -1.611 & 0.107 \\
 & Chart & Dtiuie x Seedance & 0.362 & 0.954 & 0.379 & 0.705 \\
 & Chart & Gpt x Seedance & -0.950 & 0.765 & -1.242 & 0.214 \\
 & Chart & Kling x Seedance & 0.285 & 0.634 & 0.449 & 0.654 \\
 & Chart & Osmosis x Seedance & -1.276 & 0.724 & -1.761 & 0.078 \\
 & Chart & Seathru x Seedance & 3.838 & 0.835 & 4.599 & $<0.001$ \\
 & Chart & Semi-uir x Seedance & -1.675 & 0.879 & -1.904 & 0.057 \\
 & Chart & Slurpp x Seedance & 0.406 & 0.647 & 0.628 & 0.530 \\
 & Chart & U-shape x Seedance & 1.878 & 0.707 & 2.659 & 0.008 \\
 & Chart & Qwen x Sora & -1.289 & 0.520 & -2.478 & 0.013 \\
 & Chart & Seedream x Sora & -0.994 & 0.518 & -1.920 & 0.055 \\
 & Chart & Dtiuie x Sora & 5.164 & 0.817 & 6.324 & $<0.001$ \\
 & Chart & Gpt x Sora & -0.758 & 0.585 & -1.295 & 0.195 \\
 & Chart & Kling x Sora & -0.991 & 0.520 & -1.905 & 0.057 \\
 & Chart & Osmosis x Sora & -1.519 & 0.538 & -2.821 & 0.005 \\
 & Chart & Seathru x Sora & -3.138 & 0.556 & -5.646 & $<0.001$ \\
 & Chart & Semi-uir x Sora & 0.035 & 0.665 & 0.052 & 0.959 \\
 & Chart & Slurpp x Sora & -0.071 & 0.519 & -0.138 & 0.891 \\
 & Chart & U-shape x Sora & 3.301 & 0.603 & 5.478 & $<0.001$ \\
\bottomrule
\end{longtable}
\endgroup

\twocolumn

\clearpage
\section*{Supplementary Note 8}
\begin{figure}[h]
    \centering
    \includegraphics[width=0.95\linewidth]{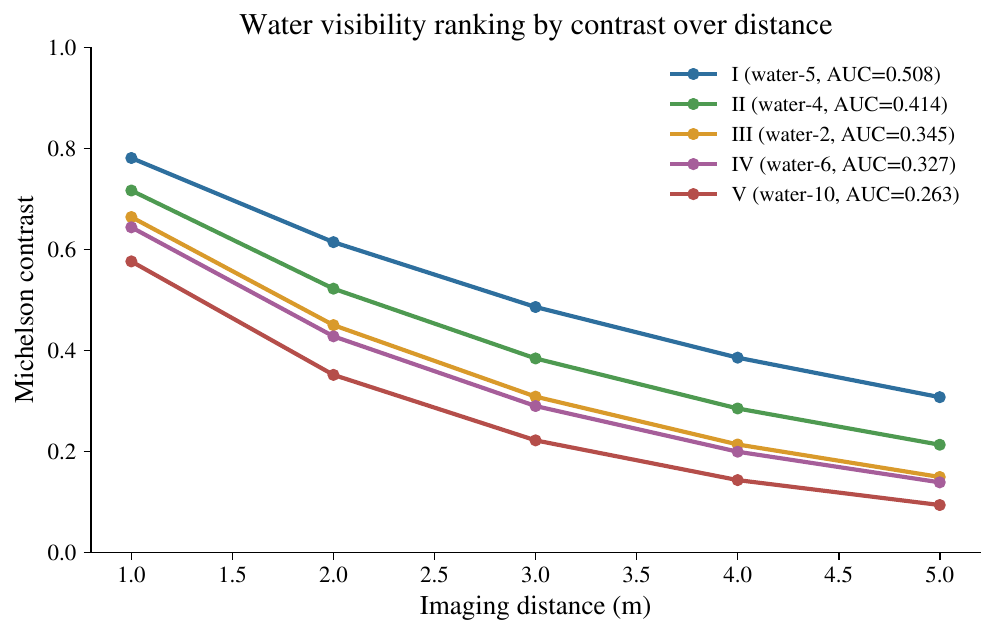}
    \caption{Michelson contrast decay curves used to rank simulated water visibility. Each curve shows contrast for a white disk target rendered at 1-5 m through one water type; legend AUC values summarize visibility across distance, with higher AUC indicating clearer water.}
    \label{fig:water_vis_curves}
\end{figure}

\textbf{Water turbidity test.}
It renders a controlled white disk on a black background through each simulated water type across distances of 1--5 m, measures luminance contrast between the disk and local background, and ranks waters by the area under the Michelson contrast-versus-distance curve (Fig.~\ref{fig:water_vis_curves}). Higher contrast indicates lower turbidity/easier visibility. This produces the easiest-to-hardest order (waters 5, 4, 2, 6, 10 in randomized-water-types database in simulator~\cite{stsr_uwsim}) which is reported in the figures as water types I, II, III, IV, and V, respectively.

\clearpage
\section*{Supplementary Note 9}
\textbf{Quantitative evaluation.} Table ~\ref{tab:real_colorchart_results} reports $\Delta E$-based color chart fidelity on real underwater images. Kling obtains the best overall performance, achieving the lowest mean, median, and standard deviation of $\Delta E$ among all compared methods. In contrast, GPT and Gemini produce similar but clearly higher errors, with mean $\Delta E$ values near 22. Slurpp, SeaThru, and Osmosis perform substantially worse and remain closer to the degraded input. Overall, the real-image color-chart evaluation supports the same conclusion as the synthetic benchmark: VLM-based restoration methods outperform conventional underwater restoration approaches, with Kling showing the strongest and most stable color recovery.

\begin{table}[H]
\centering
\begin{tabular}{lccc}
\hline
Source & Mean $\Delta E$ & Std $\Delta E$ & Median $\Delta E$ \\
\hline
Kling   & 16.128 & 10.483 & 15.051 \\
GPT     & 21.803 & 11.021 & 21.338 \\
Gemini  & 21.928 & 12.670 & 18.694 \\
Slurpp  & 26.421 & 14.367 & 22.287 \\
SeaThru & 28.630 & 14.094 & 25.638 \\
Osmosis & 33.201 & 16.973 & 27.661 \\
\hline
Raw underwater input & 34.396 & 17.505 & 28.663 \\
\hline
\end{tabular}
\caption{Summary of $\Delta E$ results on real underwater image color charts. Lower is better. The last row reports the raw underwater input and is included only for reference, not as a restoration method.}
\label{tab:real_colorchart_results}
\end{table}

\clearpage
\section*{Supplementary Note 10}
\begin{figure*}[]
    \centering
    \includegraphics[width=\textwidth]{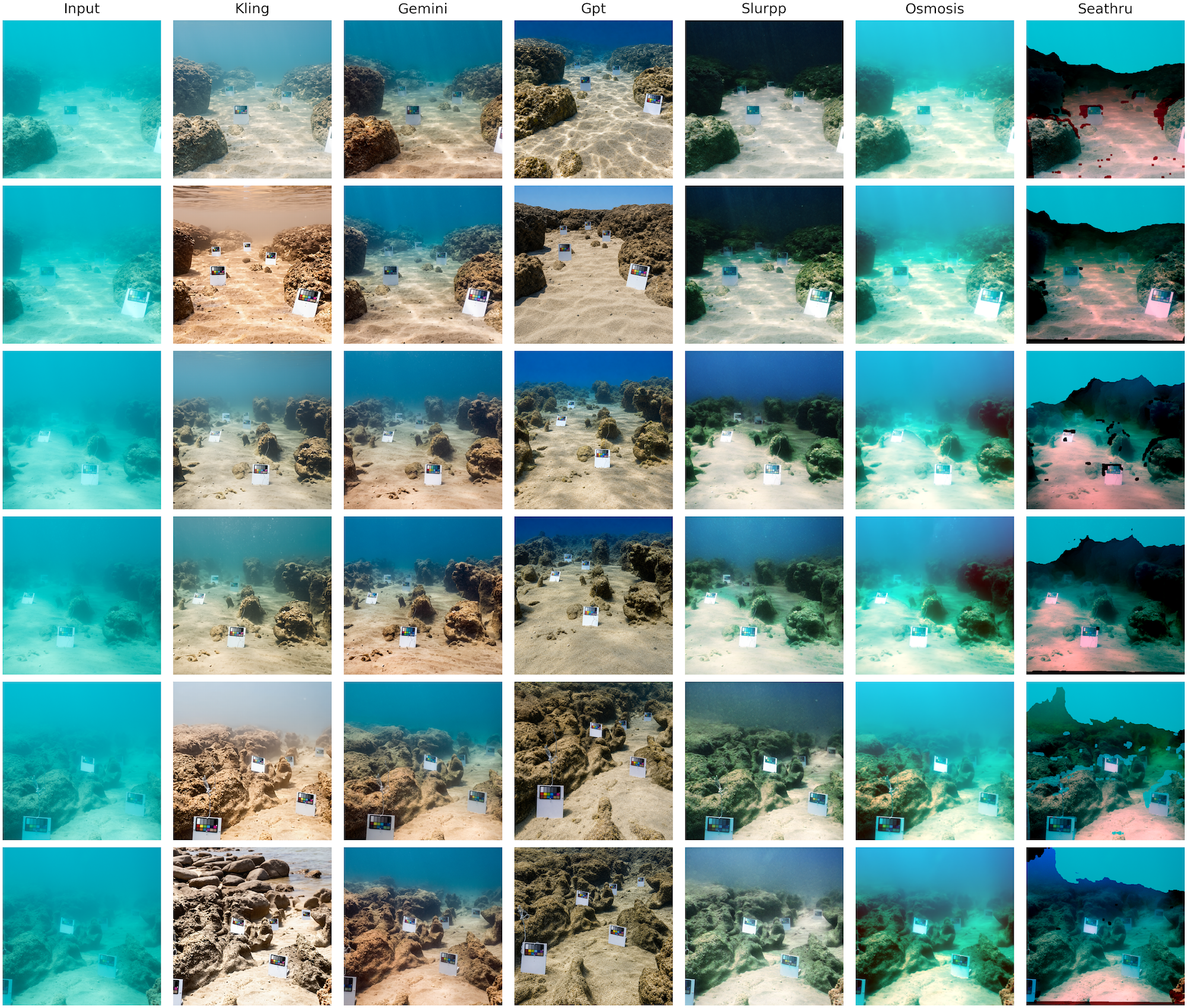}
    \caption{Additional real underwater image restoration results from Squid dataset~\cite{squid_dataset}}
    \label{fig:real_restorations_1}
\end{figure*}

\begin{figure*}[]
    \centering
    \includegraphics[width=\textwidth]{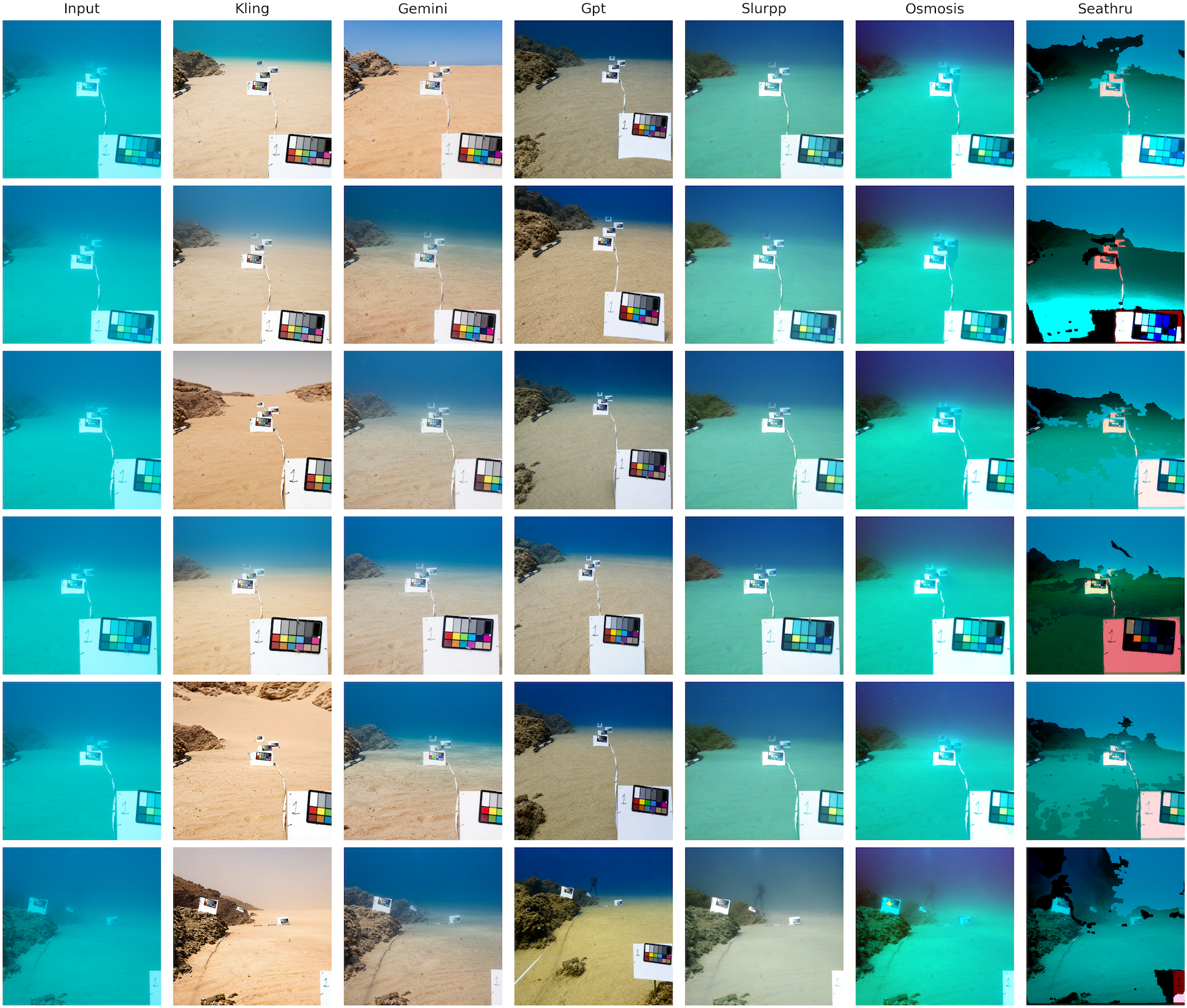}
    \caption{Additional real underwater image restoration results from Squid dataset~\cite{squid_dataset}}
    \label{fig:real_restorations_2}
\end{figure*}

\begin{figure*}[]
    \centering
    \includegraphics[width=\textwidth]{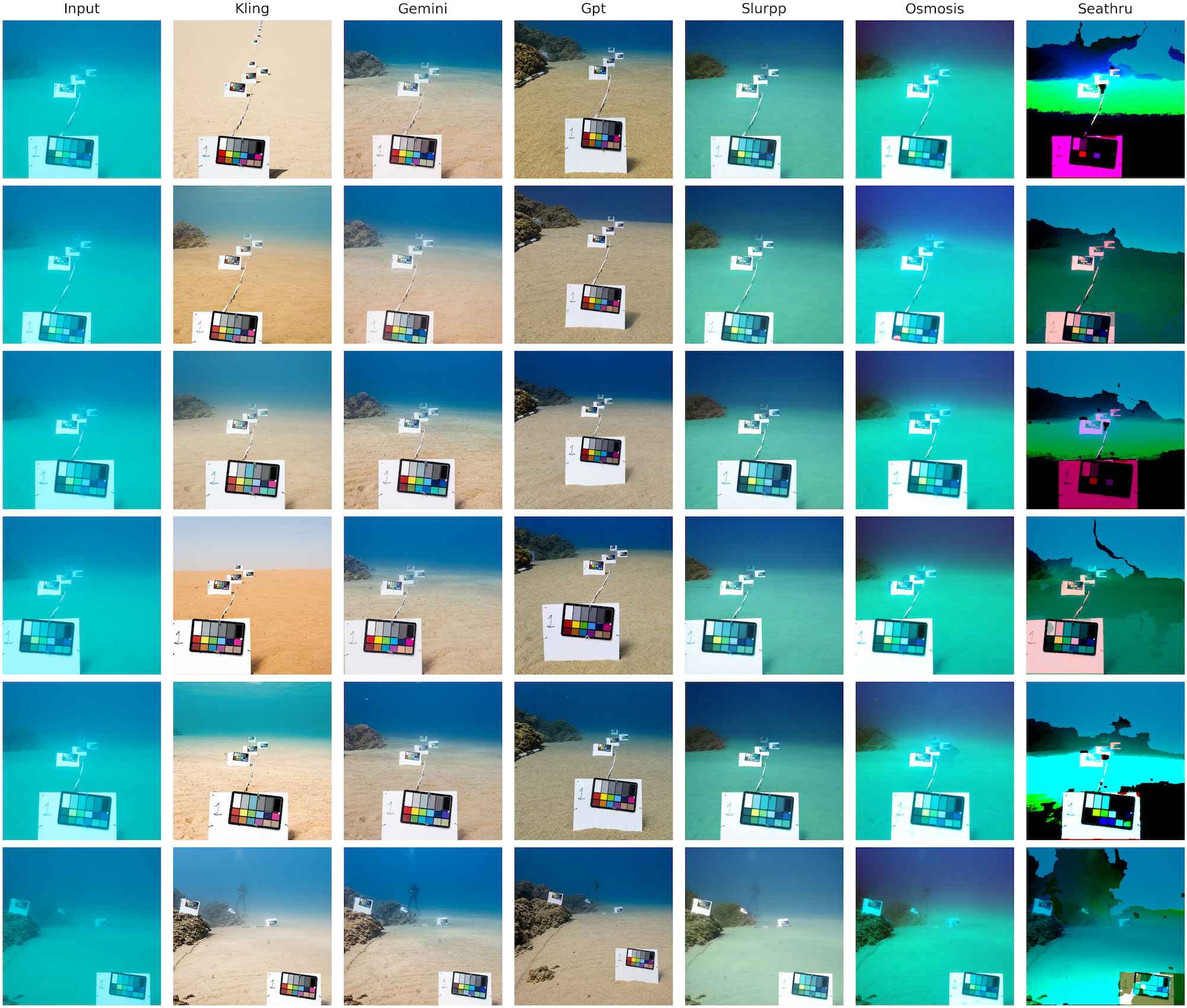}
    \caption{Additional real underwater image restoration results from Squid dataset~\cite{squid_dataset}}
    \label{fig:real_restorations_3}
\end{figure*}

\textbf{Additional qualitative results.} Real underwater image restorations from Squid dataset~\cite{squid_dataset} shown in Figs.~\ref{fig:real_restorations_1}-~\ref{fig:real_restorations_3}.

\section*{Supplementary Note 11}
\begin{figure*}[t]
    \centering

    \begin{subfigure}[t]{\textwidth}
        \centering
        \includegraphics[width=\linewidth]{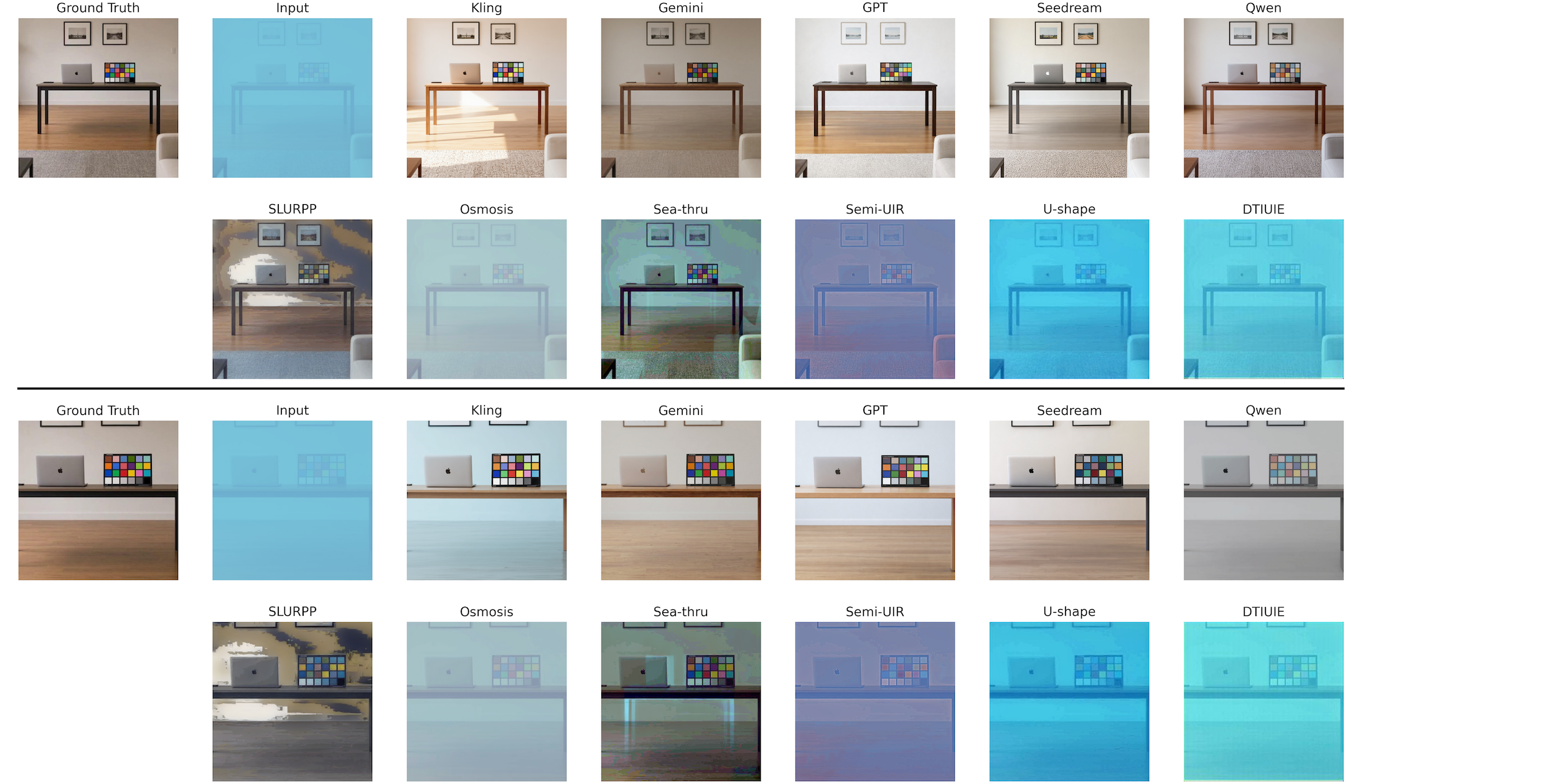}
        \caption{With easiest water I simulation.}
        \label{fig:officedesk_w5}
    \end{subfigure}

    \vspace{0.6em}

    \begin{subfigure}[t]{\textwidth}
        \centering
        \includegraphics[width=\linewidth]{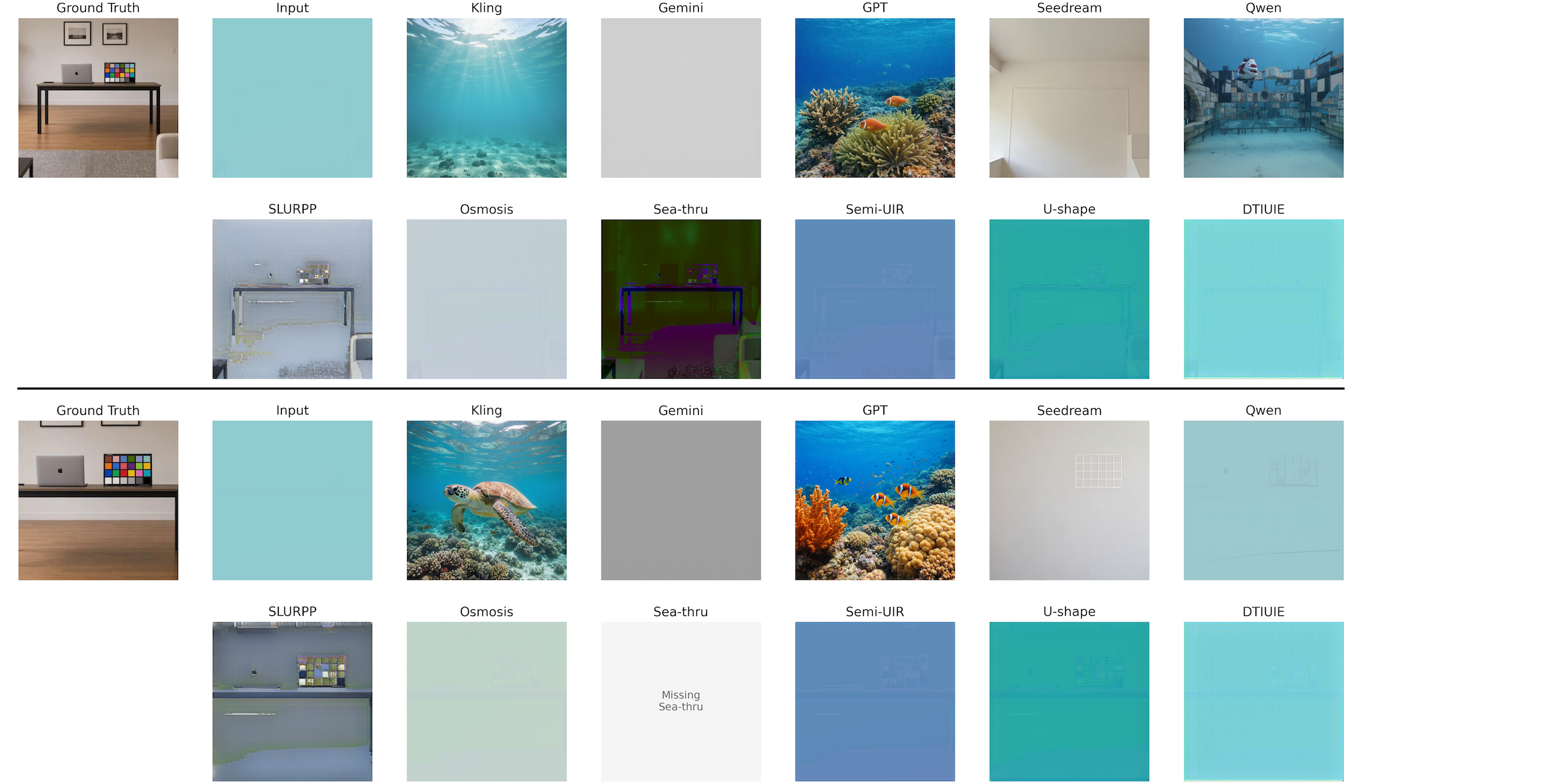}
        \caption{With most challenging water V simulation. No method successfully restores; VLMs produce irrelevant content; UIR methods fail.}
        \label{fig:officedesk_w10}
    \end{subfigure}

    \caption{Qualitative restoration comparisons are shown for a far and a close frame of the office scene in Sora-set.}
    \label{fig:officedesk-sora}
\end{figure*}

\begin{figure*}[t]
    \centering

    \begin{subfigure}[t]{\textwidth}
        \centering
        \includegraphics[width=\linewidth]{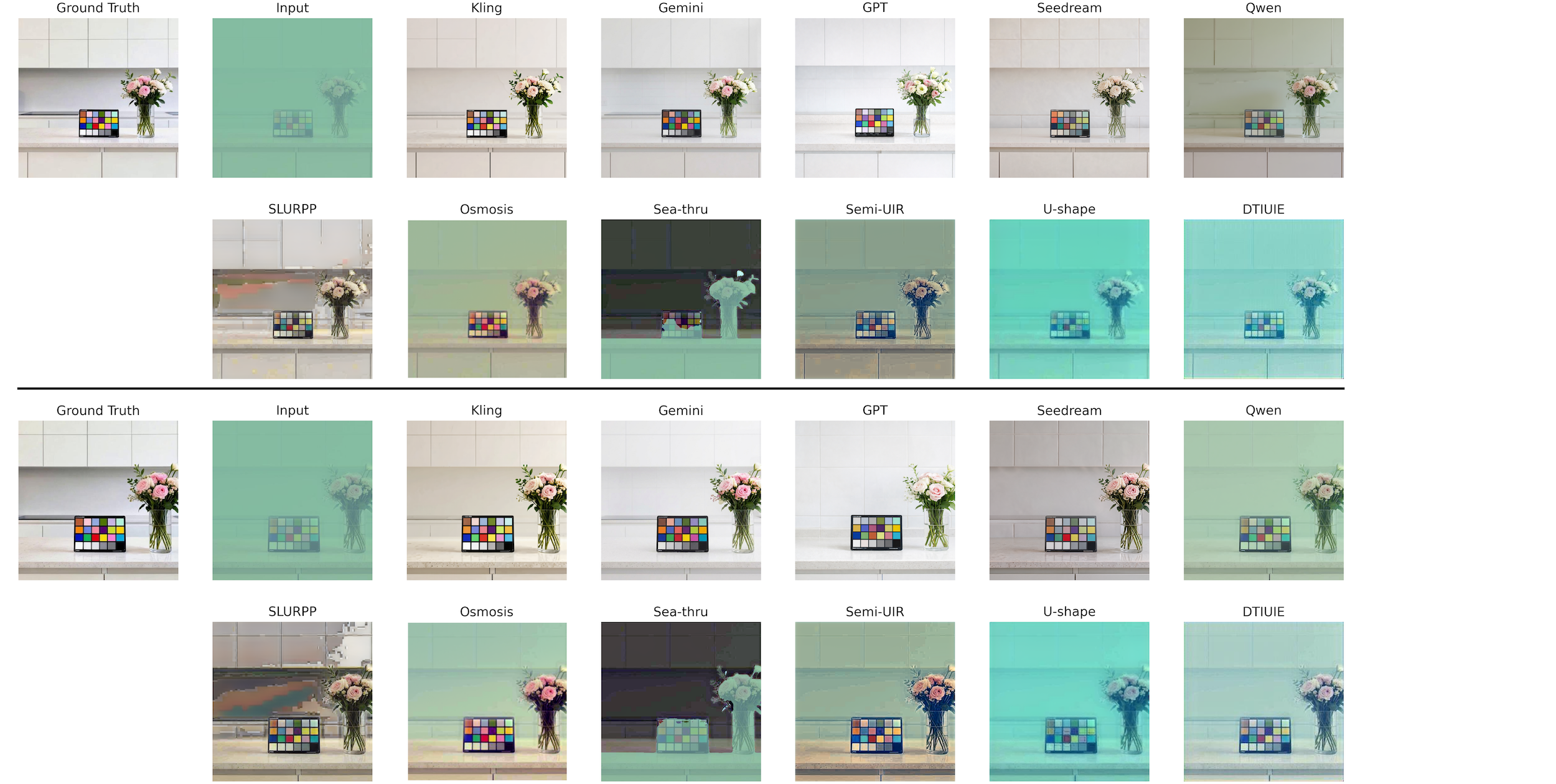}
        \caption{With water III simulation.}
        \label{fig:kitchen_w2}
    \end{subfigure}

    \vspace{0.6em}

    \begin{subfigure}[t]{\textwidth}
        \centering
        \includegraphics[width=\linewidth]{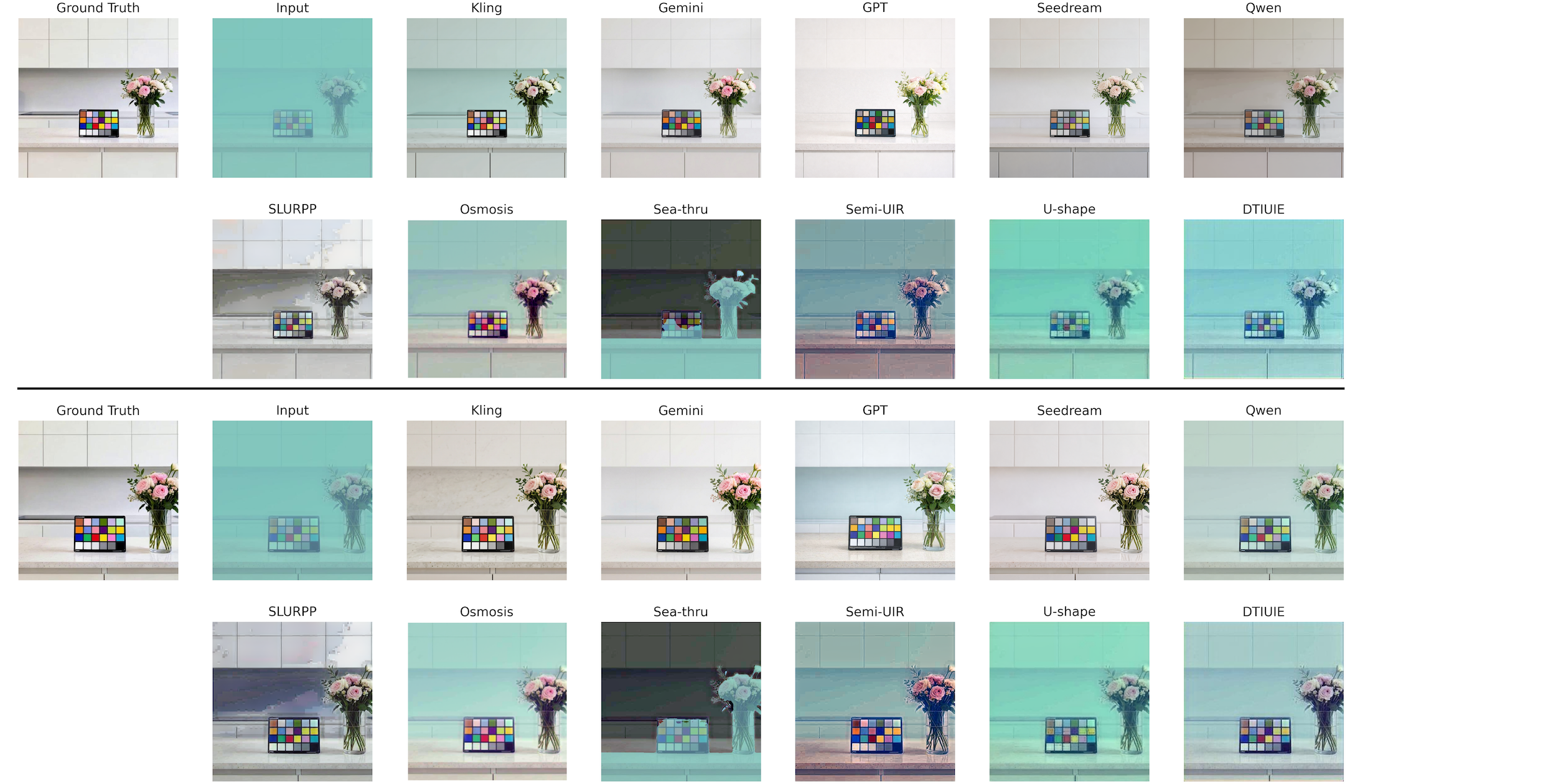}
        \caption{With water II simulation.}
        \label{fig:kitchen_w4}
    \end{subfigure}

    \caption{Qualitative restoration comparisons are shown for a far and a close frame of the kitchen scene in Seedance-set.}
    \label{fig:kitchen-seedance}
\end{figure*}

\textbf{Qualitative comparisons on our benchmark.} Two frames (far and close) per ``home office" scene and water condition (easiest I and hardest V) are shown in Figs.~\ref{fig:officedesk_w5}-\ref{fig:officedesk_w10}; per ``kitchen" scene and mid water conditions (II-III) are shown in Figs.~\ref{fig:kitchen_w2}-~\ref{fig:kitchen_w4}.

\end{document}